\pdfoutput=1
\documentclass[11pt]{article}
\usepackage{acl}
\usepackage{times}
\usepackage{latexsym}
\usepackage[T1]{fontenc}
\usepackage[utf8]{inputenc}
\usepackage{microtype}
\usepackage{inconsolata}
\usepackage{graphicx}
\usepackage{subcaption}
\usepackage{booktabs}
\usepackage{paralist}
\usepackage{longtable}
\usepackage{amsfonts}
\usepackage{tabularx}
\usepackage{dirtytalk}
\usepackage{multirow}
\usepackage{tabularx}
\usepackage{scalefnt}

\usepackage{xspace}
\newcommand{\recent}{\textsc{Recent}\xspace}
\newcommand{\creation}{\textsc{Creation}\xspace}
\newcommand{\donation}{\textsc{Donation}\xspace}
\newcommand{\F}{F$_1$\xspace}

\newcommand\blfootnote[1]{%
  \begingroup
  \renewcommand\thefootnote{}\footnote{#1}%
  \addtocounter{footnote}{-1}%
  \endgroup
}

\renewcommand{\paragraph}[1]{\noindent\textbf{#1}}

\usepackage{tabularray}
\usepackage{float}
\usepackage{rotating}
\usepackage[normalem]{ulem}
\UseTblrLibrary{booktabs}
\UseTblrLibrary{siunitx}

\NewTableCommand{\tinytableDefineColor}[3]{\definecolor{#1}{#2}{#3}}
\UseTblrLibrary{varwidth}
\SetTblrInner[tblr]{measure=vbox}

\title{Donate or Create? Comparing Data Collection Strategies for\\ Emotion-labeled Multimodal Social Media Posts}

\author{Christopher Bagdon$^{1}$, Aidan Combs$^{1,2,3}$, Carina Silberer$^{3}$, \and Roman Klinger$^{1}$\\
  $^{1}$Fundamentals of Natural Language Processing, University of Bamberg, Germany\\
  $^{2}$Department of Sociology, The Ohio State University, USA\\
  $^{3}$Institut f\"ur Maschinelle Sprachverarbeitung, University of Stuttgart, Germany\\
  \texttt{\{christopher.bagdon,roman.klinger\}@uni-bamberg.de}\\
  \texttt{combs.494@osu.edu};
  \texttt{carina.silberer@ims.uni-stuttgart.de}}

\begin{document}
\maketitle
\begin{abstract}
  Accurate modeling of subjective phenomena such as emotion expression
  requires data annotated with authors' intentions. Commonly such data 
  is collected by asking study participants to
  \textit{donate} and label genuine content produced in the real
  world, or \textit{create} content fitting particular
  labels during the study. Asking participants to create
  content is often simpler to implement and presents fewer risks to
  participant privacy than data donation. However, it is unclear if and how
  study-created content may differ from genuine content, and how
  differences may impact models. We
  collect study-created and genuine multimodal social media posts labeled
  for emotion and compare them on several dimensions, including model
  performance. We find that compared to genuine posts, study-created
  posts are longer, rely more on their text and less on their images
  for emotion expression, and focus more on emotion-prototypical
  events. The samples of participants willing to donate versus create
  posts are demographically different. Study-created data is valuable to train 
  models that generalize well to genuine data, but realistic effectiveness
  estimates require genuine data. 

\end{abstract}
\blfootnote{The work has been conducted at the University of Bamberg
and the University of Stuttgart. The Ohio State University is the new
affiliation for Aidan Combs.}

\section{Introduction}
Emotions play a fundamental role in communication
\citep{emotion_socialmedia, CHUNG2020103108}, 
particularly in online settings \citep{DERKS2008766}.
On contemporary social
media sites, authors often express emotion through a combination of text
and visual content
\citep{illendula2019multimodal,image_stats}. Modeling the emotions
expressed in social media posts therefore requires multimodal datasets
labeled for author emotion. This is, however, a challenging task:
Emotions are internal psychological states and external annotators can
therefore only approximate the correct labels \citep{Troiano2023,
  nakagawaExpressionsCausingDifferences2022}.

One approach to mitigate this annotator--author label mismatch is to
ask study participants to create content fitting provided
labels \cite[``Write a text that caused emotion
X'',]{Troiano2023}. While this is simple to implement, the resulting
data may differ from real social media data. Such lack of
generalizability may lead to limited model robustness
\citep{degtiar2023,
  elangovan-etal-2024-principles,ribeiro-etal-2020-beyond,yang-etal-2023-distribution}.

An alternative approach is to ask social media users to donate and
label their real social media posts
\citep{opreaISarcasmDatasetIntended2020}.  While this approach may
provide more realistic data, it requires more precautions to
protect participant privacy
\citep{keuschYouHaveTwo2024,gomezortegaDataTransactionsFramework2023}.

Little is known about what precisely the differences between study-created and 
donated author-labeled content may be, or how significant differences may be 
for modeling. We provide a better understanding of the tradeoffs of these 
corpus collection methods with the goal of informing future author-labeled 
corpora collection efforts. To do this, we collect study-created and genuine multimodal social 
media posts labeled by their authors for emotion. We analyze differences 
between the events that inspire the posts, how they are labeled for emotion, 
and sample characteristics. Finally, we explore the impact of these differences 
on emotion modeling and prediction.

We implement three collection procedures:
\begin{compactenum}
\item \creation: Study-created data. Participants are asked to create
  posts about an event they experienced that elicited an emotion for
  which we prompt. This approach is clear-cut to conduct but
  potentially lacks generalizability.
\item \donation: Genuine data. Participants provide posts from their
  social media accounts about an event they experienced that elicited
  a prompted emotion. This method yields real-world data balanced
  across emotions, but might come with privacy issues and
  participant's self filtering, as well as potentially limited
  availability of posts.
\item \recent: Genuine data. Participants submit their five most
  recent posts and then annotate each for emotion. While this method
  avoids potential experiment bias in emotion annotation, it may
  underrepresent emotions that are rarely shared on social media.
\end{compactenum}

We find that (1)~study-created data differs from genuine data in several
ways. Notably, it is dominated by prototypical emotion triggers, while
genuine data is more diverse.  (2)~The data collection procedures lead
to different samples of participants.  (3)~Models trained on \creation
generalize well to genuine data, but \donation test data is required
to realistically estimate their effectiveness.

\section{Related Work}
\label{sec:relatedwork}
Asking authors to label text is useful 
in areas where author intent is both important and unclear from the text 
alone. For example, author-annotated corpora exist for deception detection 
\citep{capuozzoDecOpMultilingualMultidomain2020,
  velutharambathCanFactualStatements20}, sarcasm detection \citep
{opreaISarcasmDatasetIntended2020,abufarhaSemEval2022Task62022}, and, of 
particular interest to us, emotion detection \citep
{kajiwaraWRIMENewDataset2021,troiano-etal-2019-crowdsourcing,scherer1997isear,
  Troiano2023}. In this section, we review common methods for collecting annotations of authors' internal states.

\subsection{Genuine Data Collection}
The standard approach in natural language processing and computer vision is to 
acquire genuine data from the world and request annotations from
external annotators. These annotators may be trained experts
or recruited through crowdsourcing platforms. However, since annotators 
do not have access to the original author's internal state, their annotations 
are often inaccurate \citep{kajiwaraWRIMENewDataset2021,Troiano2023,
  nakagawaExpressionsCausingDifferences2022}.

To circumvent this issue, researchers may try to
indirectly acquire labels from authors. For example, in social media data,
author labels can sometimes be inferred using corresponding hashtags
\citep{mohammadEmotionalTweets2012,abbesDAICTDialectalArabic2020}. While such 
approaches are useful when reliable markers are available, they are also 
subject to error: Annotating texts with hashtags as an indicator of a
particular label makes other such markers redundant.

If author intent is difficult to access in an indirect, rule-based
manner, an alternative is to ask social media users to donate and
label their own data \citep{opreaISarcasmDatasetIntended2020,
  kajiwaraWRIMENewDataset2021,razi2022,pfiffnerDataDonationModule2024}. Directly
obtaining labels from authors eliminates incorrect inferences as
sources of error.

Collecting genuine social media data annotated by authors is
challenging \citep{vanDriel2022}. This is because of privacy concerns
\citep{boeschoten2022,carriereBestPracticesStudies2024,gomezortegaDataTransactionsFramework2023},
the required technical skill set of participants
\citep{keuschYouHaveTwo2024}, and the potentially limited availability
of requested types of data.

\subsection{Study-created Data Collection}
An alternative to collecting real data is to construct author-labeled
datasets by asking participants to role-play or write according to
specific instructions that provide target labels. For example,
participants may be asked to tell truths or lies
\citep{ottNegativeDeceptiveOpinion2013,capuozzoDecOpMultilingualMultidomain2020,velutharambathCanFactualStatements20,lloydMiamiUniversityDeception2019},
write about events that elicited certain emotions
\citep{troiano-etal-2019-crowdsourcing,Troiano2023}, respond with
specific coping strategies \citep{troiano-etal-2024-dealing}, or act
out hypothetical emotional scenarios
\citep{busso2008iemocap,busso2016msp}.

Such approaches for obtaining study-created data have various
advantages. They avoid the difficulties of finding a sufficient amount
of content fitting uncommon labels ``in the wild''.  
Participants have more control over what they contribute, and can better guard their privacy. 
Finally, the data can be less prone to recall bias as large time
gaps between content creation and labeling are avoided.

\begin{figure}[t]
  \centering
  \includegraphics[width={.8\columnwidth}]{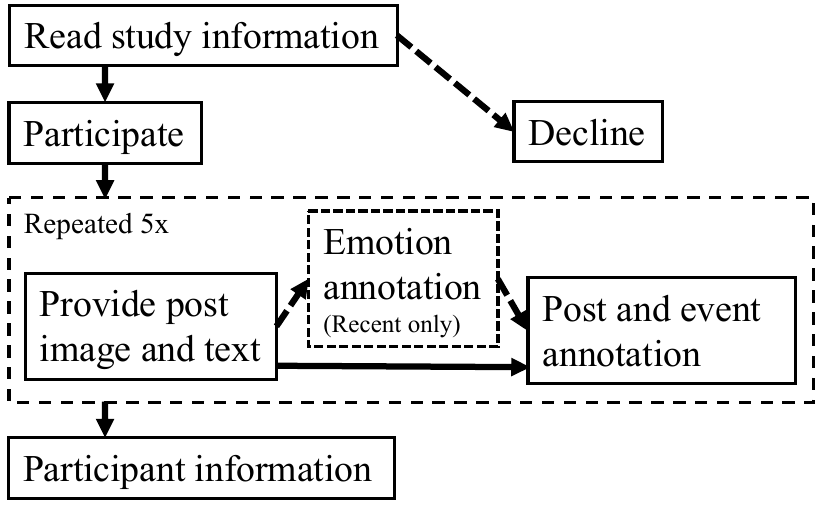}
  \caption{Data collection process.}
  \label{fig:flowchart}
\end{figure}

\begin{table*}[t]
  \small
  \renewcommand{\arraystretch}{0.8}
  \begin{tabularx}{\linewidth}{p{3cm}Xl}
    \toprule
    Label & Question Text & Options \\
    \cmidrule(r){1-1}\cmidrule(rl){2-2}\cmidrule(l){3-3}
    \multicolumn{3}{l}{\textbf{Emotion annotation} (\recent only)}   \vspace{.8ex}  \\
    Emotion & Please select the emotions that you felt as a result of this
              event [multiple] & [Emo.]
    \\
    Intensity & Please rate how intensely you felt each of these emotions
                as a result of this event [Emo.] & 1\ldots5
    \\
    \cmidrule(r){1-1}\cmidrule(rl){2-2}\cmidrule(l){3-3}
    \multicolumn{3}{l}{\textbf{Text--image relationship:}  How much do these statements apply? }   \vspace{.8ex} \\
    Text describes image & The text directly describes the image.  &
                                                                     1\ldots5
    \\
    Text $\rightarrow$ image & The text is required to understand
                               the image.  & 1\ldots5
    \\
    Image $\rightarrow$ text & The image is required to understand
                               the text.  & 1\ldots5
    \\
    Image conveys emotion & The image explicitly conveys the emotion you
                            posted about.  & 1\ldots5
    \\
    Text conveys emotion & The text explicitly conveys the emotion you
                           posted about.  & 1\ldots5
    \\
    \cmidrule(r){1-1}\cmidrule(rl){2-2}\cmidrule(l){3-3}
    \multicolumn{3}{l}{\textbf{Event experience}}  \vspace{.8ex} \\
    Event Duration & How long did the event last?  & [Time]
    \\
    Emotion Duration & How long did you experience emotion as a result
                       of the event?  & [Time]
    \\
    Event Intensity & How intense was your experience of the event?  &
                                                                       1\ldots5
    \\
    Emotion Intensity & How intense was your experience of this emotion?
                          & 1\ldots5
    \\
    \cmidrule(r){1-1}\cmidrule(rl){2-2}\cmidrule(l){3-3}
    \multicolumn{3}{p{14cm}}{\textbf{Appraisal:} Think back to when the event happened and recall its details. Take some time to remember it properly.
    How much do these statements apply? Some statements might not fit the event exactly, please answer to the best you can.}   \vspace{.8ex} 
    \\
    Familiarity & The event was familar. & 1\ldots5\\
    Predictability & I could have predicted the occurrence of the event. & 1\ldots5\\
    Attention & I had to pay attention to the situation. & 1\ldots5\\
    Notconsider & I tried to shut the situation out of my mind. & 1\ldots5\\
    Pleasantness & The event was pleasant for me. & 1\ldots5\\
    Unpleasantness & The event was unpleasant for me. & 1\ldots5\\
    Goalrelevance & I expected the event to have important consequences for me. & 1\ldots5\\
    Ownresponsibility & The event was caused by my own behavior. & 1\ldots5\\
    Goalsupport & I expected positive consequences for me. & 1\ldots5\\
    Anticipconseq & I anticipated the consequences of the event. & 1\ldots5\\
    Owncontrol & I was able to influence what was occurring during the event. & 1\ldots5\\
    Otherscontrol & Someone other than me was influencing what was occuring. & 1\ldots5\\
    Acceptconseq & I anticipated that I would easily live with the unavoidable consequences of the event. & 1\ldots5\\
    Effort & The situation required me a great deal of energy to deal with it. & 1\ldots5\\
    Internalstandards & The event clashed with my standards and ideals. & 1\ldots5\\
    \bottomrule
  \end{tabularx}
  \caption{Wording and response options for survey questions used in the analysis. [Emo.] refers to Anger, Disgust, Fear, Joy, Sadness,
    Surprise. [Time] refers to one of Seconds, Minutes,
    Days, Weeks, Months.}
  \label{tab:postquestions}
\end{table*}

Such methods may, however, be susceptible to other experiment effects
\citep{vaniaAskingCrowdworkersWrite2020}. For instance, people behave
differently when they know they are participating in a study, often to
confirm the hypotheses they think the researchers have
\citep{MUMMOLO_PETERSON_2019,Nichols2008}. More emotionally intense
events are remembered more easily, potentially leading participants to
preferentially select them when writing posts about them
after-the-fact \citep{kensinger_remembering_2009}. Writing differs
based on audience, for example varying in the degree to which it is
stereotype-consistent
\citep{lyons2006,lyonsHowAreStereotypes2003a}. Differences between
study-created data and actual social media content are likely to harm the
generalizability of models trained on study-created data \citep[see][for a
discussion of differences between hold-out data and real-world
data]{elangovan-etal-2024-principles,ribeiro-etal-2020-beyond,yang-etal-2023-distribution}.

\section{Data Acquisition Methods}
\label{sec:methods}
In this section we discuss the data acquisition methods and
their advantages and disadvantages.\footnote{All data, code, and 
  surveys are available at \url{https://www.uni-bamberg.de/en/nlproc/projects/item/}}

\subsection{Process Overview}
The data collection process is illustrated in
Figure~\ref{fig:flowchart}. Potential participants see the study
details and choose to participate or decline. Those who accept are
asked to provide a social media post using one of three data
collection strategies -- \creation, \donation, and \recent{} -- and
then annotate it. Each participant provides five posts. 
Finally,
participants answer questions about themselves. Participants may 
complete the study as many times as they wish, up to once per emotion.

\subsection{Data Collection Strategies}
\label{sec:datacollection}
We request posts that contain both text and an image and are authored
by the participant. Other instructions vary by collection strategy.

\paragraph{\creation.}
We ask participants to recall an event in which they felt a particular
emotion -- anger, disgust, joy, fear, sadness, or surprise -- and
which they remember well. They then write a social media post about
it. Participants must select an image from the Flickr database of
Creative Commons licensed
images\footnote{\scalebox{0.99}[1]{\url{https://www.flickr.com/creativecommons/by-2.0/}}}
that is similar to the one they would have used if they had created
the post naturally. This approach diminishes the risk to participant
privacy, but posts may differ from genuine data.

\paragraph{\donation.}
We prompt participants for an emotion and ask them to share a
genuine post from their timeline. They do so by copy--pasting the text
and uploading the image from that post. This approach yields genuine
posts while equally representing emotions regardless of their prevalence on 
social media. However, participants may struggle to find posts representing uncommon target emotions. This approach also raises greater privacy
concerns.\footnote{Participants are informed of potential privacy
  risks, and consent to them before beginning the study.}

\paragraph{\recent.}
\recent is similar to \donation, but we do not prompt participants for
particular emotions. Instead, we ask them to share their five most
recent multimodal posts and annotate each with all emotions, and the associated intensities, they felt in response to the event that inspired the post. We
adapt this emotion annotation approach from \citet{rhodes2021measuring}.\footnote{We did not ask 
  participants to declare a primary emotion in the case of ties for highest intensity. We further discuss this
  in the Limitations section.}
\recent diminishes concerns about the accuracy of emotion annotations at the 
cost of potentially underrepresenting uncommon emotions. The same privacy 
concerns that affect \donation also apply to \recent.

\subsection{Annotation Details}

\paragraph{Post and Event Annotations.} Table~\ref{tab:postquestions}
shows the questions participants answer about each of their posts. To
understand the roles of images and text, we ask about the relationship
between the modalities. To understand the link between the event and
the emotion category, we request appraisal labels
\citep{scarantino2016philosophy,scherer}. Appraisals are a psychological theory 
of how events induce emotion \citep{Troiano2023,Stranisci2022}.

\paragraph{Participant Information.} After participants annotate their
posts, we ask about their age, gender, education, ethnicity, frequency
of social media use and posting, and choice of social
media platform. We additionally use demographic information provided
by 
the research platform we use -- namely, employment and student status. We summarize
these questions in Table~\ref{tab:participantinformation} in the
Appendix.

\paragraph{Study Details.}
We recruit participants using Prolific. We require that they reside in
the United Kingdom or Ireland, have resided there for at least five
years, are~18 or older, and be native English speakers. We restrict
participation to these demographics to avoid confounding variables, as 
emotion expression can differ from culture to culture. The survey is conducted via Google Forms. 
\creation and \donation take 30~minutes and we pay participants~£4.50. \recent takes 40~minutes and we pay participants~£6.00.
After removing posts that do not meet our requirements, our dataset
contains 2,507~posts authored by 522~participants.

\section{Analysis and Modeling}
\label{sec:analysis}

We answer research questions about differences in the data
between collection strategies (\textbf{RQ1}), differences in the
events that lead to the posts (\textbf{RQ2}), differences between
samples (\textbf{RQ3}) and in how participants assign labels
(\textbf{RQ4}), and the impact on model performance and generalization
(\textbf{RQ5}).

\begin{table}[t]
  \centering\small
  \begin{tabular}{l rr rr rr}
    \toprule
    & \multicolumn{6}{c}{Study} \\
    \cmidrule(lr){2-7}
    Emotion & \multicolumn{2}{c}{\creation} & \multicolumn{2}{c}{\donation} & \multicolumn{2}{c}{\recent} \\
    \cmidrule(r){1-1}\cmidrule(r){2-3}\cmidrule(rl){4-5}\cmidrule(l){6-7}
    Anger    &  193 & 17\% &  174 & 15\% &  16 & 8\%  \\
    Disgust  &  202 & 17\% &  195 & 17\% &  10 & 5\%  \\
    Fear     &  193 & 17\% &  189 & 16\% &   4 & 2\%  \\
    Joy      &  189 & 16\% &  196 & 17\% & 158 & 79\% \\
    Sadness  &  185 & 16\% &  197 & 17\% &  16 & 8\%  \\
    Surprise &  197 & 17\% &  198 & 17\% &  30 & 15\% \\
    \cmidrule(r){1-1}\cmidrule(r){2-3}\cmidrule(rl){4-5}\cmidrule(l){6-7}
    Total    & 1159 &       & 1149 &       & 199 &       \\
    \bottomrule
  \end{tabular}
  \caption{Distribution of posts in the dataset. In \recent, posts are
    categorized into emotions based on which emotion the participant
    reported feeling most intensely. In the case of ties, the post is
    counted in both categories.}
  \label{tab:CorpusContents}
\end{table}

\subsection{RQ1: Are there differences in posts between data collection strategies?}
\label{subsection:rq1}
Table~\ref{tab:CorpusContents} shows the post and label
distributions. We see that \creation and \donation represent emotions
nearly evenly by design. In \recent, we obtain posts dominated by joy.

\begin{figure}
  \centering
  \includegraphics[width=\columnwidth]{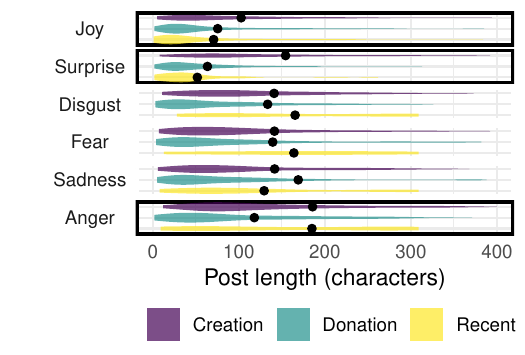}
  \caption{Length of the posts, in characters, by emotion and
    study. Means are represented by points. Outlined boxes
    indicate significant differences (one-way ANOVA, p<0.05
    after Bonferroni correction).}
  \label{fig:postlength}
\end{figure}

\paragraph{Text.} Figure~\ref{fig:postlength} shows the average lengths in
character counts. \mbox{\creation} posts are longer than \donation and
\recent posts, especially for posts about joy and surprise. Controlling for differences in emotion distribution, \creation
posts are 51\% longer than \recent posts and 26\% longer than
\donation posts (p<0.01 for both).\footnote{Examples can be seen in
  Appendix \ref{app:example_posts}.}

\paragraph{Image style.}
To investigate differences in image style, we manually label each
image as a \textsl{meme, screenshot, graphic, professional photo,
  personal photo,} or \textsl{other}.\footnote{Details and examples are in Appendix~\ref{sec:Appendix_analysis}.} Figure
\ref{fig:image_styles} shows the distribution of image styles across the
three data collection strategies. They differ
significantly between studies ($\chi^2$ test p<0.001). Personal photos
are most prominent across studies. The remaining images
for \creation are dominated by
professional photos. \donation and \recent have fewer
professional photos and instead more screenshots, graphics, memes, and
other images, which are less prevalent in the study-created data.

\begin{figure}[t]
  \centering
  \begin{subfigure}[t]{0.45\columnwidth}
    \centering
    \includegraphics[width=\linewidth]{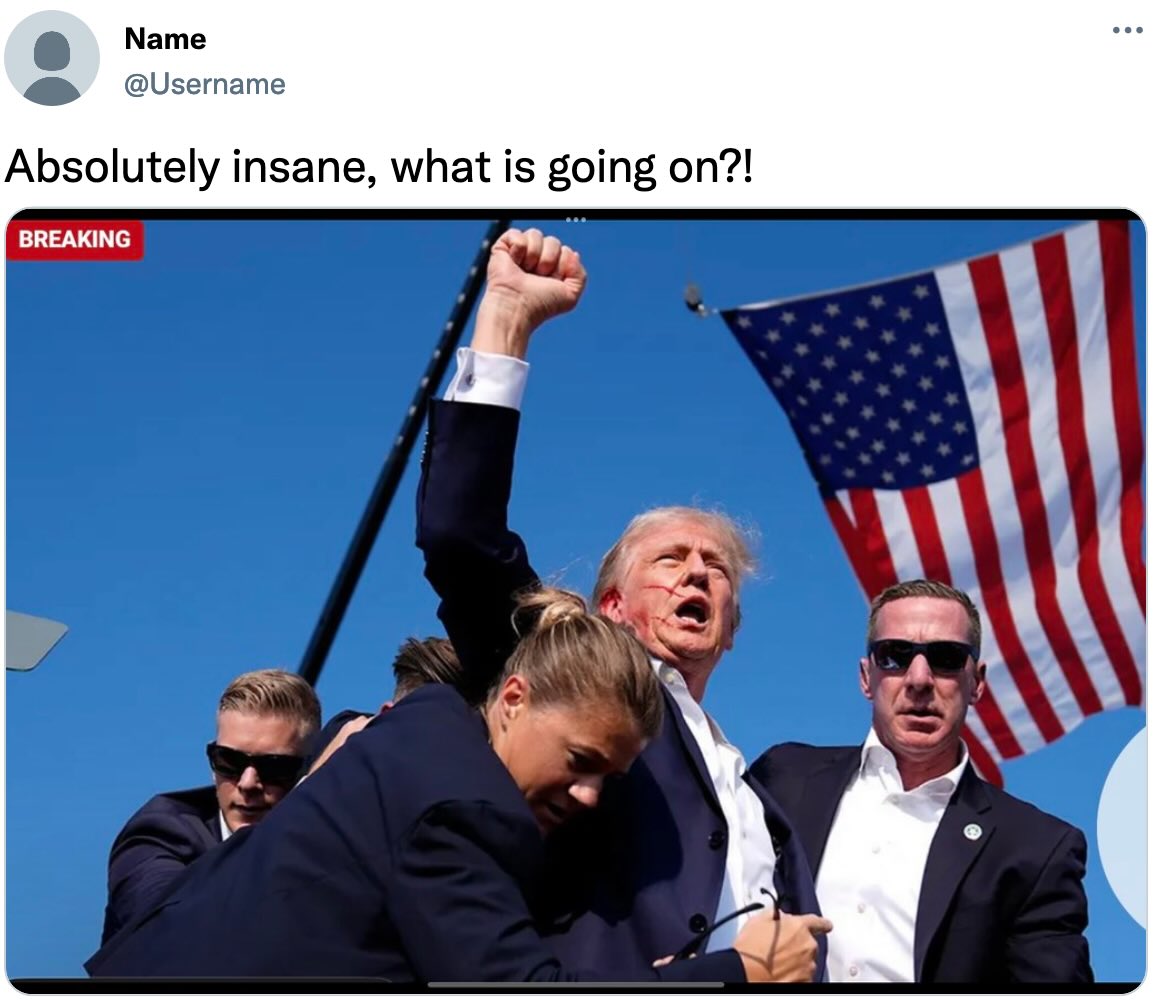}
    \caption{\creation\ post labeled as surprise.}
    \label{fig:image1}
  \end{subfigure}
  \hfill
  \begin{subfigure}[t]{0.45\columnwidth}
    \centering
    \includegraphics[width=\linewidth]{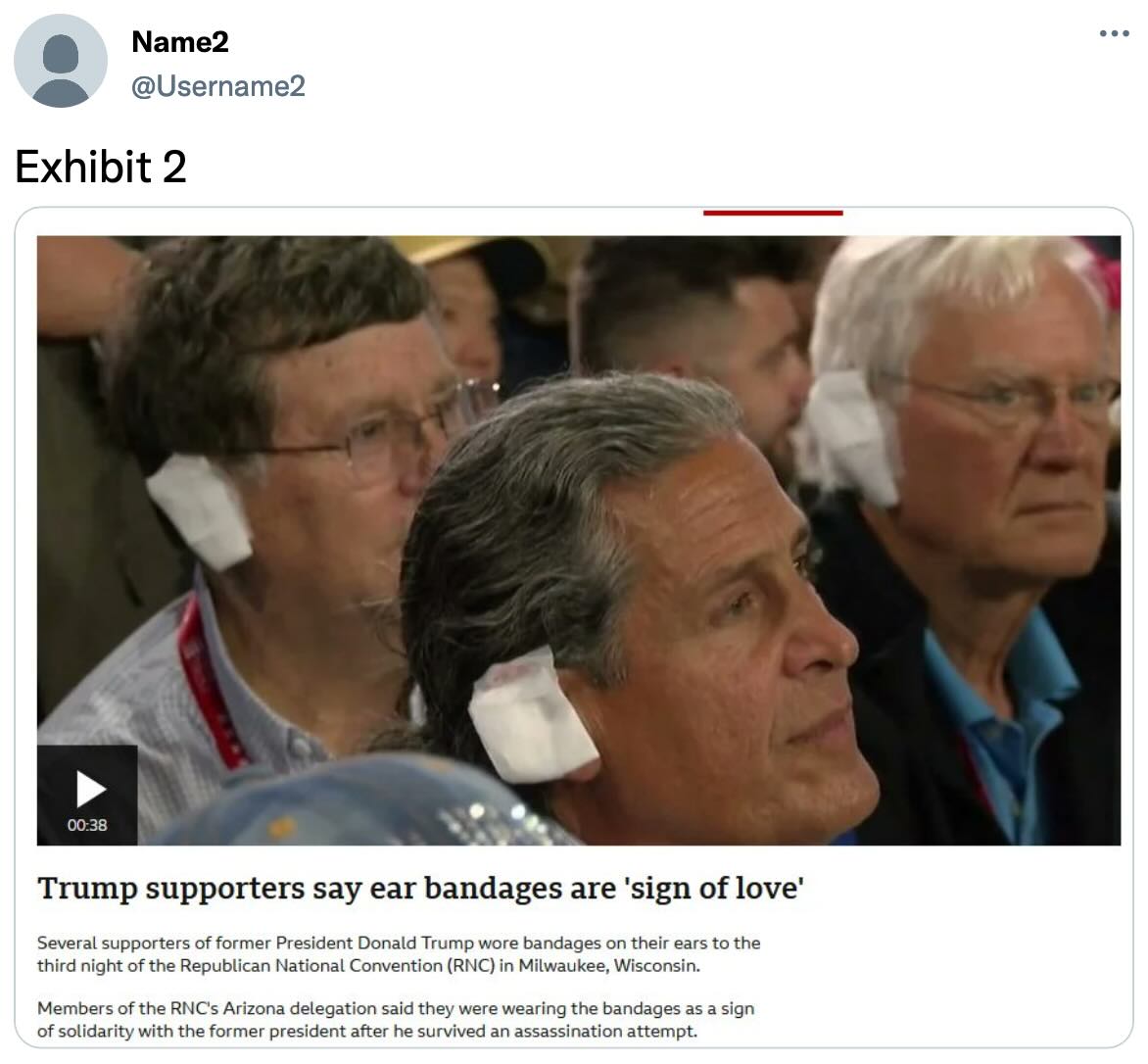}
    \caption{\recent\ post labeled as anger.}
    \label{fig:image2}
  \end{subfigure}
  \caption{Comparison of posts submitted for \creation\ and \recent. \creation\ is a more general comment, while \recent\ is a reaction to a specific news item. The latter style does not occur in the \creation\ data.}
  \label{Figure:Event_diff}
\end{figure}

The lack of screenshots in \creation is a consequence of requiring
participants to select images from an existing database.
Figure~\ref{Figure:Event_diff} illustrates this difference. This is
backed by feedback from study participants who found the database
limiting. Further, it shows that study-created data might not include all
real-world triggers for social media posts.
\begin{figure}[t]
  \centering
  \includegraphics[width=\columnwidth]{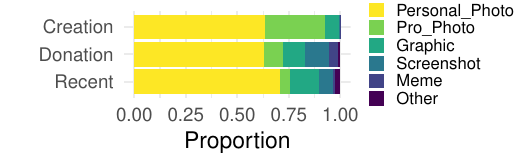}
  \caption{Distribution of image style labels.}
  \label{fig:image_styles}
\end{figure}

\begin{table}[t]
  \centering\small
  \setlength{\tabcolsep}{3pt}
  \begin{tabular}{lr lr lr}
    \toprule
    \multicolumn{2}{l}{Creation} & \multicolumn{2}{l}{Donation} &\multicolumn{2}{l}{Recent} \\
    \cmidrule(r){1-2}\cmidrule(lr){3-4}\cmidrule(lr){5-6}
    text             & 43\%      & text             & 55\%      & text            & 58\%     \\
    background       & 12\%      & tree             & 11\%      & tree            & 14\%     \\
    tree             & 12\%      & background       & 9\%       & woman           & 13\%     \\
    sky              & 11\%      & woman            & 9\%       & man             & 12\%     \\
    grass            & 11\%      & sky              & 9\%       & sky             & 12\%     \\
    building         & 9\%       & person           & 8\%       & background  & 11\%     \\
    water            & 8\%       & people           & 8\%       & person          & 10\%     \\
    people           & 8\%       & grass            & 8\%       & grass           & 9\%      \\
    woman            & 7\%       & man              & 8\%       & smile           & 8\%      \\
    car              & 7\%       & smile            & 7\%       & building        & 8\%        \\ \bottomrule
  \end{tabular}
  \caption{10 most common image content words per study. Image content
    word extraction done by GPT-4o.}
  \label{Table:Image_content}
\end{table}

\begin{figure*}[t]
  \centering
  \includegraphics[width=\linewidth]{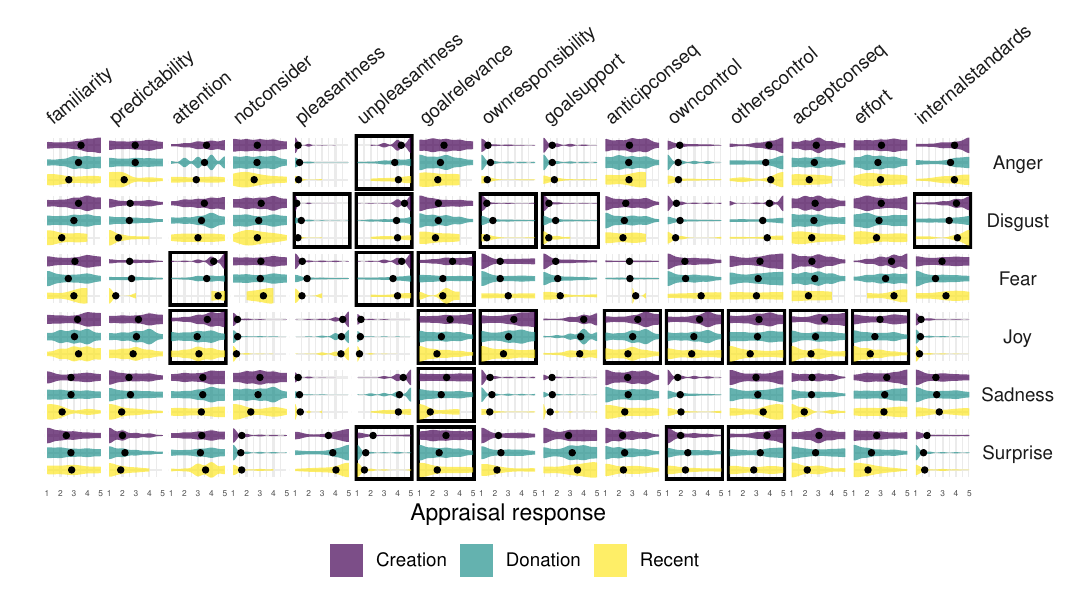}
  \caption{Responses to event appraisal questions (columns)
    across emotions (rows) and studies (color). Responses were
    provided on 5-point Likert scales. Means are represented by
    points. Outline: Significant at p<0.05 according to an ANOVA with
    Bonferroni correction.}
  \label{fig:appraisal}
\end{figure*}
\paragraph{Image content.}
We label the images with GPT-4o to analyze the
content.\footnote{\url{https://openai.com/gpt-4}, prompt and validation
  details are in Appendix~\ref{sec:Appendix_analysis}.} Table~\ref{Table:Image_content} shows the 10 most
frequent labels for each study (Table~\ref{tab:top_50} in the Appendix
shows the 50 most frequent labels). Despite the clear differences in image style that we observed, the word lists indicate that the images show comparable content across all collection strategies, but with differences in the order. The most common for
all three was text, though this was less frequent for \creation (43\%) than for
\donation (55\%) and \recent (58\%). This is consistent with the smaller number of screenshots and lack of memes in \creation.  
The image labels are dominated by
references to people or scenery.

\begin{figure}[t]
  \centering
  \includegraphics[width=\columnwidth]{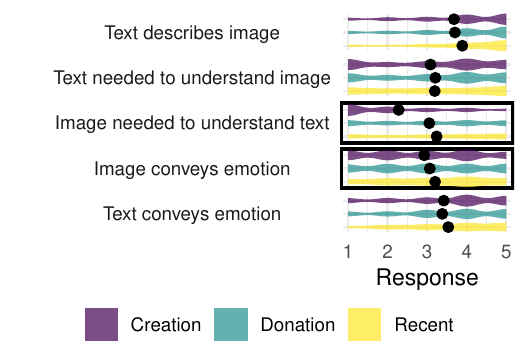}
  \caption{Participant ratings of the relationship between post text
    and images, on five point Likert scales. Means are represented
    by points. Outline indicates significance (one-way ANOVA, p<0.05
    after Bonferroni correction).}
  \label{fig:textimrelation}
\end{figure}

\paragraph{Text--Image Relation.}
To understand the role of the image and the text in conveying an
emotion, we asked participants questions about the relationship
between the two.  Figure \ref{fig:textimrelation} shows the response
distributions. \creation participants
describe their post images as less necessary to understand the text
and as conveying emotion less than \donation and \recent
posts. \creation post images are not as integral to the post as a
whole. In contrast, in \recent posts, the text and images are more
dependent on one another.

\subsection{RQ2: Are there differences in the events that inspire participants to write posts?}
\label{subsection:rq2}
With this and the following research questions, we aim at understanding potential reasons
for the differences that we observe in the posts.

\paragraph{Emotion Appraisals.}
Figure~\ref{fig:appraisal} shows the distribution of responses for the
event emotion appraisal survey questions by study and post emotion. On
22 of the 90~question--emotion combinations (15~questions $\times$ 6~post emotions), we find statistically significant differences between
studies. Joy shows significant differences in more questions (8 of 15~questions) than any other emotion.\footnote{The reason is, presumably,
  that the imbalanced corpus in \recent leads to more power to detect significant
  differences for Joy. When \recent is excluded from consideration, there are
  similarly many significant differences between \creation and
  \donation for Joy as for the other emotions.}

In most cases, the events that inspired study-created posts are rated more
highly on the appraisal dimensions than the events that inspire
genuine posts. The cases which break this pattern tend to be those
appraisal dimensions which are negatively associated with the emotion
at hand. For example, participants rate disgust-causing events as less
pleasant and less likely to have positive consequences in
\creation. \recent posts, in contrast, tend to rate lowest on these
appraisal dimensions, particularly in joy and surprise, where they are
best represented. This suggests that participants use more
prototypical events for particular emotions in \creation than in
genuine posts.

\paragraph{Duration and Intensity.}
Events that are recalled by prompting for a specific emotion may be
dominated by the emotion and the duration and intensity of the
event. To analyze this, we estimate mixed-effects models that include
the study type and emotion as independent variables and random
intercepts to account for grouping by participant.  We find that, on a
5-point Likert scale, controlling for emotion, participants rate
\creation events as 0.34 points more intense than \donation events,
and their emotional responses to \creation events as 0.36 points more
intense (p < 0.001 for both). There are no intensity differences
between \donation and \recent events or emotional responses.

Participants report that emotion responses to \creation events last
significantly longer than responses to \donation events (p < 0.001),
and responses to \donation events last significantly longer than
responses to \recent events (p < 0.01). There are no significant
differences between \creation and \donation in the length of the
events themselves. However, \recent events are significantly shorter
than \donation events (p < 0.01).

It is possible that these observations are a consequence of the role that 
emotions play in memorizing events. In \creation, participants may preferentially select events that are more memorable and more emotionally prototypical than those they actually post about on social media. 
\begin{figure}[t]
  \noindent
  \centering
  \includegraphics[width=\columnwidth]{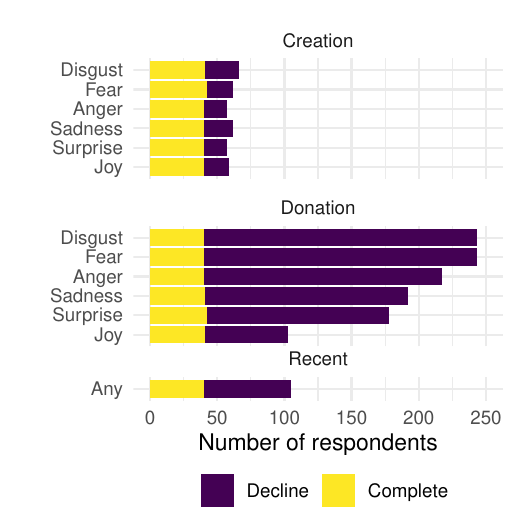}
  \caption{Differences in decline rates between studies and
    emotions. ``Decline'' means that a participant opted not to
    participate after reading the study information.}
  \label{fig:returnrates}
\end{figure}
\subsection{RQ3: Are there differences in participant characteristics?}
\label{subsection:rq3}
We find that participants who completed \creation are significantly
older, less likely to be students, and more likely to be European than
those who completed \donation and \recent.\footnote{Table~\ref{tab:sampledemographics} in the Appendix shows the sample
  demographics.}

\recent and \donation require that participants be comfortable sharing
their real social media posts with researchers, and, additionally,
\donation requires that those posts be about specific emotions. These
differences in study requirements may change who is willing to
participate. Figure~\ref{fig:returnrates} reports the decline rates
after having read the instructions. Decline rates are considerably higher in
\donation and \recent than in \creation ($\chi^2$ test, p < 0.001).
Participants are indeed hesitant to provide researchers with
their social media posts.

In \donation we find differences between decline rates across emotions
($\chi^2$ test p < 0.001). Participants finish the study more often
when asked for joy-inducing event posts. The fact that the \creation
decline rate does not vary by emotion suggests that people do not find it
particularly challenging to create posts about emotions other than
joy, but rather they struggle to find appropriate posts in their
social media feeds. Participants who declined \donation and \recent
commonly cited privacy concerns and, in \donation, they noted a lack of posts
that reflected the desired emotion.

\subsection{RQ4: Are there differences in how participants label posts for emotion?} 
\label{subsection:RQ4}
Some emotions are harder to find on social media, and some posts may be about 
events which evoke more than one emotion. This may lead \donation participants 
searching for uncommon emotions to submit posts which also or better represent 
common emotions. In such cases, the labels of the posts would be affected by 
the target labels presented to the participants, which is an undesirable 
potential source of bias. To investigate the potential scope of this issue, we 
allowed participants to label their posts freely with any number of emotions in 
\recent. We presume that the annotations of multi-emotion posts are likely to 
be more affected by the target label presented to the participant.

\begin{figure}[t]
  \centering
  \includegraphics[width=\columnwidth]{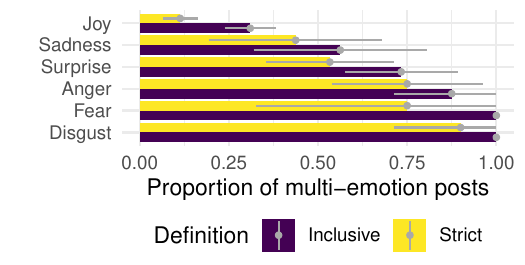}
  \caption{\protect Prevalence of multi-emotion posts in the \recent study. Lines represent 95\% confidence intervals. 
    Posts were counted as multi-emotion under the Inclusive definition if the participant indicated the event evoked two or more emotions, and under the Strict definition if the participant indicated that two or more emotions were tied for greatest intensity.}
  \label{fig:multiemo}
\end{figure}

Figure \ref{fig:multiemo} shows the fractions of posts to which multiple 
emotions were assigned. Multi-emotion posts are common. Most anger, fear 
disgust, and surprise posts were additionally labeled with another emotion. We 
conclude that bias in \donation emotion annotations is a possibility that
should be investigated and mitigated. One approach would be to allow 
participants to
label their posts with multiple emotion labels and the intensity of
those emotions, as we did in \recent.

\subsection{RQ5: Do data differences affect model performance?}
\label{sec:modeling}
\label{subsection:rq5}
Our experiments aim to assess if \creation and \donation are equally
suitable as training and as testing data for predictive
models\footnote{We do not train or test on \recent because of its smaller size and unbalanced emotion distribution.}. Specifically, do data differences between \creation and
\donation have an effect on model performance when used as training
data? Are there differences in effectiveness when testing on
\creation vs. \donation? We fine-tune unimodal and multimodal
models separately on \creation and \donation data subsets. We also evaluate multimodal foundation models in a zero-shot
setup.

\paragraph{Setup.}
We divide the data such that the development and test sets each have
25 posts per emotion (300 posts per set) and use the remaining data
for training (800 posts per strategy).\footnote{\recent is not included
  in the test set for reasons detailed in the Limitations Section.} 
As unimodal models, we use
RoBERTa for text \citep{liu2019roberta} and ViT for images
\citep{dosovitskiy2021an}. For multimodal models, we use the dual
encoder CLIP \cite{pmlr-v139-radford21a}, applying early fusion by
concatenating text and image embeddings, then adding a classification
head on top. Appendix~\ref{app:modeldetails} gives model and training details. 
All models are fine-tuned and tested five times. 

For the zero-shot setup, we prompt
llama3.2-vision,
llava-llama3,
and minicpm-v.
We prompt each model five times and report results averaged across
runs. We select the best model per modality on the development data
and report the corresponding result on the test data\footnote{Three
  other zero-shot models were also tested, but returned results that
  were either worse in all instances or unparseable. See
  Appendix~\ref{app:modeldetails} for details on these models.}. We
provide model details and prompts in Appendix~\ref{app:modeldetails}.

\begin{table}
  \centering\small
  \begin{tabular}{llc cc c}
    \toprule
    &&&\multicolumn{2}{c}{Training} \\
    \cmidrule(lr){4-5}
    && Mod. & \donation & \creation & Zero-shot \\
    \cmidrule(r){2-3}\cmidrule(lr){4-4}\cmidrule(lr){5-5}\cmidrule(lr){6-6}
    \multirow{6}{*}{\rotatebox{90}{Test \F}} 
    &\multirow{3}{*}{\rotatebox{90}{Creatn}} 
      & V  & .16 & .18 & .24\textsuperscript{1} \\
    && T  & .49 & .58  & .61\textsuperscript{1} \\
    && T+V & .60 & .62 & .56\textsuperscript{2} \\
    \cmidrule(r){2-3}\cmidrule(lr){4-4}\cmidrule(lr){5-5}\cmidrule(lr){6-6}
    &\multirow{3}{*}{\rotatebox{90}{Donatn}} 
      & V  & .19 & .18  & .19\textsuperscript{3}  \\
    && T  & .41 & .42  & .45\textsuperscript{1}  \\
    && T+V & .38 & .40 & .43\textsuperscript{2} \\
    \bottomrule
  \end{tabular}
  \caption{Performance of models predicting emotion in multimodal
    social media posts using text alone (T), image alone (V), and both
    modalities combined (T+V). We report macro \F scores over 5
    runs. Zero-shot models are chosen according to the best
    individual performance on development
    data. \textsuperscript{1}llama3.2-vision. \textsuperscript{2}minicpm-v. \textsuperscript{3}llava-llama3.}
  \label{tab:results}
\end{table}

\paragraph{Results.}
Table~\ref{tab:results} shows the main results.\footnote{Full results
  including precision, recall, and \F-scores for individual emotions are in
  Appendix~\ref{app:modeling}, Tables~\ref{tab:zeroshotapp} and
  \ref{app:modelingapdx}.}  Models trained on \creation and 
\donation data perform equally well when tested on \donation data (bottom block in Table~\ref{tab:results}). 
This suggests that the differences between \creation and 
\donation content may not be very important for model training. Performance
scores on \creation test data are higher (top block), and likely unrealistically optimistic, in comparison
to scores on \donation test data. This suggests that genuine data
is required to reliably estimate model effectiveness. The zero-shot models' 
results, which by design solely reflect differences in the test sets, underpin
this finding.

\paragraph{Influence of post and respondent features.}
Our analyses in Sections~\ref{subsection:rq1}, \ref{subsection:rq2},
\ref{subsection:rq3}, and~\ref{subsection:RQ4} show that collection
strategies lead to differences in samples, and our analyses show that
these differences carry over to model performances. We now investigate
this relationship between post and respondent features and model
performance further.

We fit three separate logistic regression models to test for
predictors of correct emotion classification with three multimodal
models whose performance is shown in Table \ref{tab:results}. CLIP
trained on \donation and \creation, and minicpm-v as a zero-shot
approach. Our dependent variable is a binary variable indicating
whether the model predicted the emotion in a given test set post
correctly on a particular run. To account for non-independence between
predictions for the same post from each of the five runs, we use
clustered standard errors.

Independent variables are text length, image style, and text--image
relation variables (cf.\ RQ1), emotion appraisals and event and
emotion duration and intensity (RQ2), and participant
sociodemographics (RQ3). Numeric and ordinal variables are scaled to
$[0;1]$. For categorical variables, we choose reference
levels (which serve as our baseline for each variable) following
\citet{johfre2021reconsidering}.\footnote{Refer to
  Appendix~\ref{app:regressiontab} for additional details.} We control
for the study the post is from and the emotion it shows. We
omit 27 posts for which at least one independent variable
is missing.

\begin{figure}[t]
  \centering
  \includegraphics[width=\columnwidth]{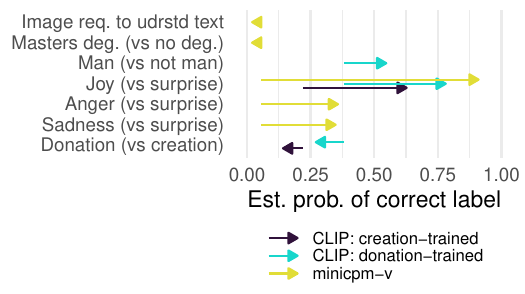}
  \caption{\protect Influence of post and respondent features on T+V model accuracy. Arrows represent predicted change in the probability of a correct post classification when the indicated variable is changed from its reference value (categorical variables) or from its minimum to its maximum (numeric variables). Only effects significant at the p<.05 level or better are shown.}
  \label{fig:regression}
\end{figure}

Figure \ref{fig:regression} shows the effects of the statistically significant variables on the probability that a post will be classified correctly.\footnote{Full results are reported in Appendix Table \ref{app:regressiontab}.} Arrows begin at the probability of correct classification indicated by the model intercept: that is, when all independent variables are either 0 (numeric/ordinal) or at their reference values (categorical). Arrow heads indicate the change in this probability when the indicated independent variable is changed, either to its maximum value (numeric/ordinal) or to the indicated category (categorical).

Minicpm-v is less likely to predict emotion accurately when authors report the image is necessary to understand the post text. Figure \ref{fig:textimrelation} shows that \creation posts have images that are less necessary for text understanding. Together, these two results suggest performance on \creation data may be better in part because \creation posts have images that are less integral to text interpretation.

Minicpm-v accuracy is also lower for posts authored by people with advanced degrees compared to those who have not gone to college. The CLIP model trained on \donation is more likely to predict emotion correctly for posts authored by men. Notably, these effects are net of the effects of emotion and all other variables in the regression. These sociodemographic effects underscore the importance of being attentive to sample composition.

Consistent with our results above, Figure \ref{fig:regression} shows that some emotions, and particularly Joy, are predicted more accurately than others. CLIP is less likely to accurately predict emotions for \donation posts as compared to \creation posts. That this effect is significant despite the presence of the other independent and control variables indicates there are further study differences that are not accounted for here. 

We did not observe statistically significant impacts on classification
outcomes for other variables not included in
Figure~\ref{fig:regression}. We note that our test set size is
relatively small and effects are therefore conservative. The
relationship between post/author features and model performance is
an important avenue for future research.

\section{Conclusion}
\label{sec:conclusion}
In this paper, we compared methods for collecting study-created and genuine 
multimodal social media posts labeled by their authors for emotion. Our 
work is the first to directly compare these methods of collecting
author-labeled corpora, and the first multimodal social media corpus
labeled by its authors for emotion.

Our results show that \creation posts are different in content 
and style and represent more prototypical events than genuine posts collected 
in \donation and \recent. \recent leads to a more realistic emotion 
distribution, but does not evenly represent all emotions, which is a challenge 
for model development. More participants are comfortable participating in 
\creation, suggesting that this corpus may better represent social media
users. We note that all presented approaches may be easily scaled up to 
large quantities of posts. Corpus sizes are only limited by available 
participants on study platforms such
as Prolific, and, for genuine data strategies, the amount of data 
they can provide.

Despite content differences, \creation training data leads to models that
perform on par with models trained on \donation  data.  Modeling results
on \donation test data show worse -- and likely more realistic --
performance than on the \creation test data. Therefore, we suggest
that future studies consider using a strategy similar to \donation for
developing test sets and \creation to collect corpora for model
development as needed for model optimization.

\section*{Acknowledgments}
We thank all ARR reviewers for their thorough reviews and valuable
suggestions. We also thank our study participants for contributing to
our work.  This work has been supported by the Deutsche
Forschungsgesellschaft (DFG) in the project ``User’s Choice of Images
and Text to Express Emotions in Twitter and Reddit'' (ITEM, Project KL
2869/11-1, No.\ 513384754).

\section*{Limitations}
We took precautions to allow for the best comparability of the data
that we obtained by our studies, but some pragmatic decisions were
required in the design. Most importantly, the studies were performed
sequentially, each with multiple phases. As such, the time periods in
which posts were collected differed and could potentially affect the
content of posts. For example, there are more posts about the US
presidential assassination attempt in the study which occurred days
after the event than in studies which occured weeks later. We do
however note that we do not find evidence in our analysis for such
differences that would influence the conclusions of our work. In all 
three conditions there were posts on a variety of topics, containing 
a variety of images -- posts about current events are not a 
dominant portion of our dataset. The image content analysis
and our qualitative review of posts supports. The requirement in 
donation and creation to find or create posts about specific emotions 
likely pushed respondents to reach further back in their feeds, 
mitigating recency effects to some extent. For example, participants 
submitted posts related to COVID lockdowns which occurred 3--4 years 
prior to data collection. That said, we acknowledge it as a limitation,
particularly in recent, and we encourage future collections of this type 
to bear it in mind and spread out their data collection if possible, 
as we did with recent.

Another limitation may be that we did not control for the
(hypothetical or real) social media platform.  Posts from different
social media platforms may differ in content and style and may
therefore not be directly comparable. While we do ask for the platform
a post is from, we do not control for this. We do a short analysis of
the distribution of platforms in our collected data in Appendix
\ref{app:platforms}. We consider platform effects to be an important
direction for future work.

As mentioned in Section~\ref{sec:datacollection} and Section~\ref{subsection:RQ4}, 
for the \recent study we
asked participants to label their posts for all emotions they experienced
in response to the event that inspired their post. While we do ask for emotion
intensity ratings, in the case of ties we did not ask them to select a primary emotion. 
This is a limitation of our work, as it prevents us from using \recent posts 
in our modeling analysis, as in the case of ties we cannot assign a single
emotion label to the post, making them incompatible with posts from \creation and \donation.
We suggest future work, for all collection strategies, both ask participants to label 
all emotions and intensities and to select a primary emotion in the case of ties.

The training and testing of models is limited by the size of our 
dataset. These experiments are designed to inform us the best methods
to collect data, and as such we will use the findings to expand the dataset,
allowing us to confirm our results in future work.

The focus of our modeling analysis is to compare training and test sets. As such, 
the pretrained models we use are selected based on their
established performance and reliability, rather than seeking 
state-of-the-art performance.

Furthermore, for the zero-shot models, we 
chose to only use models which we could run locally for two
reasons: (1) Reproducibility; changes to models such as GPT-4o are out
of our control. There is no guarantee that we or other researchers
will have access to the exact model used in our experiments. Furthermore,
there is no guarentee that changes to the model will be
disclosed to users. (2) Costs; When budgeting our research we
prioritize increasing data collection efforts over using LLM API
services. Especially given the above explanation about our model selection
choices.

\section*{Ethical Considerations}

This study was approved by the ethics review board at the University
of Bamberg. In all data acquisition efforts, all
participants have been informed about the collection procedure and the
use of the data. Nevertheless, we would like to reflect on
various potential challenges in this work.

While the participants have been informed about the use of the data,
it may sometimes be the case that the content of a post does comprise
anonymity, and the study participant may not be aware of the potential
impact. This is not an issue with the study-created data, but may be a
challenge in the \recent and \donation data. Further, the collected
data may contain information about other people not actively
participating in the study. Because of these two challenges, we
decided to only share the \recent and \donation data for research
purposes upon request.

We use a Creative Commons licensed image database as to best avoid
copyright issues. Copyright regulations vary widely by country. In
some countries, fair use rules allow for using images protected by
copyright for the purposes of academic research. However, in others
research use of copyrighted images may be a copyright
infraction. Given the international nature of emotion recognition
research, we wish to prioritize data collection methods which produce
data that can be used in as many legal contexts as possible. However,
teams that wish to use image data must consider their own local
context when designing data collections and using available corpora.

VS Code Copilot was used to assist in the writing of the code for the
data analysis. Copilot was only used for debugging, documentation,
refactoring, and code completion.

\bibliography{custom}

\appendix

\newpage
\onecolumn

\newcommand{\nt}{\hspace{5mm}}

\section{Appendix}
\label{sec:appendix}
\subsection{Survey Details}

We provide the detailed questions on the participant information in
Table~\ref{tab:participantinformation} and the details of the participant
information collected in Table~\ref{tab:sampledemographics}. We allow
participants to participate in each emotion/data collection strategy
combination once. We ask them to only complete this information the
first time. If they provide it more than once anyways, we use their
most recent data. For the small number of participants who did not
complete the study demographic information, we impute it using
demographic data provided by Prolific.\\[0.5\baselineskip]

\begingroup
\small
\begin{tabularx}{0.95\linewidth}{p{3cm}Xp{4cm}}
  \toprule
  Label & Question Text & Options \\
  \cmidrule(r){1-1}\cmidrule(rl){2-2}\cmidrule(l){3-3}
  \multicolumn{3}{l}{\textbf{Participant information}} \\
  \nt{}Age & How old are you? & $\mathbb{N}^{\geq18}$\\
  \nt{}Gender& With which gender(s) do you identify? [multiple] & Woman, Man, Nonbinary, Transgender, Other [write-in]  \\
  \nt{}Education & What is the highest level of education you completed? & No formal qualifications, Secondary education, High school, Undergraduate degree (BA/BSc/other), Graduate degree (MA/MSc/Mphil/other), Doctorate degree (PhD/other)  \\
  \nt{}Ethnicity & With which of the following ethnic groups do you identify the most? [multiple possible] & Australian/New Zealander, North Asian, South Asian, East Asian, Middle Eastern, European, African, North American, South American, Hispanic/Latino, Indigenous, Other [write in] \\
  \nt{}Social Media Use & Approximately how often do you use social media (browse or participate)? & Every day, 4-6 days a week, 2-3 days a week, Once per week, Occasionally, but less than once per week, Never\\
  \nt{}Social Media Post & Approximately how often do you post on social media?& Every day, 4-6 days a week, 2-3 days a week, Once per week, Occasionally, but less than once per week, Never \\
  \nt{}Preferred platform   & What is your preferred social media site?& X (Twitter), Facebook, Instagram, Reddit, Tiktok, LinkedIn, Other [write-in] \\
  \bottomrule
\end{tabularx}
\captionof{table}{\label{tab:participantinformation} Wording and response options for participant
  information.}
\endgroup
\vfill
\mbox{}

\begingroup
\small\centering
\begin{tabular}{l rr rr rr}
  \toprule
  & \multicolumn{2}{c}{Creation} & \multicolumn{2}{c}{Donation} & \multicolumn{2}{c}{Recent} \\
  \cmidrule(r){1-1}\cmidrule(lr){2-3}\cmidrule(lr){4-5}\cmidrule(l){6-7}
  \multicolumn{4}{l}{\textbf{Gender identity}}\\
  \nt{}Man                                                     & 109 & (0.46)  & 119& (0.5)   & 22& (0.56) \\
  \nt{}Woman                                                   & 126 & (0.53)  & 114& (0.48)  & 17& (0.44) \\
  \nt{}Other                                                   & 2& (0.01)    & 3& (0.01)    & 0         \\
  \multicolumn{4}{l}{\textbf{Ethnicity *}} \\
  \nt{}European                                                & 196& (0.84)  & 178& (0.76)  & 29& (0.72) \\
  \nt{}Non-European or multiethnic                 & 38& (0.16)   & 57& (0.24)   & 11& (0.28) \\
  \multicolumn{4}{l}{\textbf{Education}} \\
  \nt{}No college degree                                       & 81& (0.34)   & 73& (0.31)   & 12& (0.32) \\
  \nt{}Bachelor's degree                                       & 97 &(0.4)    & 99& (0.42)   & 15& (0.39) \\
  \nt{}Master's degree or higher                               & 63& (0.26)   & 62& (0.26)   & 11& (0.29) \\
  \multicolumn{4}{l}{\textbf{Employment status}} \\
  \nt{}Full-Time                                               & 136& (0.59)  & 140& (0.6)   & 27& (0.68) \\
  \nt{}Part-Time                                               & 40& (0.17)   & 41& (0.18)   & 7& (0.17)  \\
  \nt{}Not employed (unemployed, homemaker, retired, disabled) & 45& (0.19)   & 40& (0.17)   & 5& (0.12)  \\
  \nt{}Other                                                   & 10& (0.04)   & 12& (0.05)   & 1& (0.03)  \\
  \multicolumn{4}{l}{\textbf{Student *}} \\
  \nt{}Yes                                                     & 32& (0.14)   & 60& (0.26)   & 9& (0.22)  \\
  \nt{}No                                                      & 199& (0.86)  & 175& (0.74)  & 31& (0.78) \\
  \multicolumn{4}{l}{\textbf{Social media use}} \\
  \nt{}Every day                                               & 203& (0.85)  & 202& (0.86)  & 35 &(0.88) \\
  \nt{}2--6 days a week                                         & 30& (0.13)   & 29& (0.12)   & 3 &(0.07)  \\
  \nt{}Once per week or less                                   & 6& (0.03)    & 5& (0.02)    & 2& (0.05)  \\
  \multicolumn{4}{l}{\textbf{Social media post}} \\
  \nt{}Every day                                               & 19& (0.08)   & 32& (0.14)   & 2& (0.05)  \\
  \nt{}2--6 days a week                                         & 65& (0.27)   & 63& (0.27)   & 15 &(0.38) \\
  \nt{}Once per week or less                                   & 157& (0.65)  & 141&(0.6)   & 23 &(0.58) \\
  \multicolumn{4}{l}{\textbf{Preferred platform}} \\
  \nt{}Facebook                                                & 86& (0.35)   & 78 &(0.33)   & 16& (0.4)  \\
  \nt{}Instagram                                               & 79& (0.32)   & 78& (0.33)   & 14& (0.35) \\
  \nt{}X (Twitter)                                             & 37& (0.15)   & 52& (0.22)   & 7& (0.17)  \\
  \nt{}Other                                                   & 42& (0.17)   & 30 &(0.13)   & 3& (0.07)  \\
  \multicolumn{4}{l}{\textbf{Mean age (std. dev) *}} \\
  & 38.4& (12.3) & 33.2& (10.7) & 36& (11.8) \\
  \bottomrule
\end{tabular}
\captionof{table}{\label{tab:sampledemographics} Sample composition of the three studies. For categorical variables, values are counts and proportions are shown in parentheses. For age, values are sample means and standard deviations are shown in parentheses. Proportions sum to less than one due to small numbers of participants missing information on given variables. Starred categories are those for which a $\chi^2$ test (categorical variables) or one-way ANOVA (age) shows statistically significant study differences at at least the p<0.05 level.}
\endgroup

\subsection{Analysis Details}
\label{sec:Appendix_analysis}
\subsubsection{Image label and content analysis}
\begin{figure*}[p]
  \centering
  \begin{subfigure}{0.35\textwidth}
    \centering
    \includegraphics[width=\linewidth]{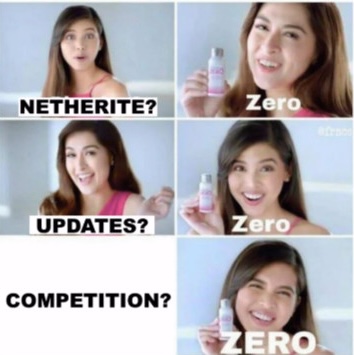}
    \caption{Meme}
  \end{subfigure}
  \hspace{5mm}
  \begin{subfigure}{0.35\textwidth}
    \centering
    \includegraphics[width=\linewidth]{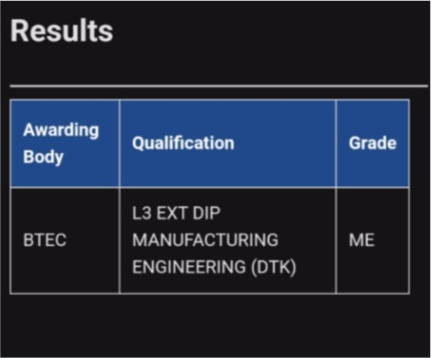}
    \caption{Screenshot}
  \end{subfigure}\\[\baselineskip]
  
  \begin{subfigure}{0.35\textwidth}
    \centering
    \includegraphics[width=\linewidth]{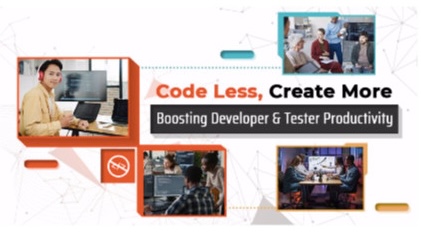}
    \caption{Graphic}
  \end{subfigure}
  \hspace{5mm}
  \begin{subfigure}{0.35\textwidth}
    \centering
    \includegraphics[width=\linewidth]{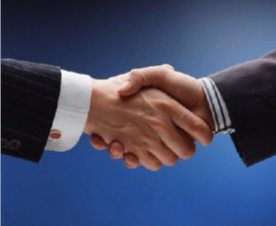}
    \caption{Professional Photo}
  \end{subfigure}\\[\baselineskip]
  
  \begin{subfigure}{0.35\textwidth}
    \centering
    \includegraphics[width=\linewidth]{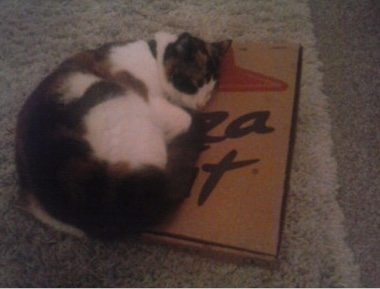}
    \caption{Personal photo}
  \end{subfigure}
  \hspace{5mm}
  \begin{subfigure}{0.35\textwidth}
    \centering
    \includegraphics[width=\linewidth]{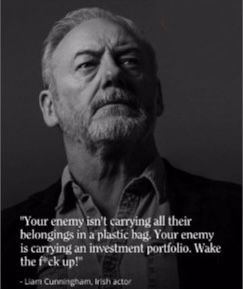}
    \caption{Other}
  \end{subfigure}
  \caption{Examples of image labels}
  \label{fig:image_label_example}
\end{figure*}

\paragraph{Image labels:}
To investigate differences in image style, we manually label each
image as a \textsl{meme, screenshot, graphic, professional photo,
  personal photo,} or \textsl{other}. We inductively developed this
list from our observations of our data. The list of labels is
hierarchical, such that if the image is a meme and a personal photo,
it will be labeled as meme. Examples of each can be seen in Figure
\ref{fig:image_label_example}. \textsl{Memes} use an established meme
format edited to the authors' specific use. \textsl{Screenshots} are
images of the authors' computer or smart phone screen. These includes
shots of programs, websites, news stories, other social media posts,
etc. \textsl{Graphic} refers to any image drawn, painted, or otherwise
created and is not a photograph, but includes photographs which have
been edited with graphic overlays. \textsl{Professional photos} are
photographs which are \textbf{clearly} a stock photo or created for
professional purposes such as news, sports media, or
advertisements. All other photographs are labeled as \textsl{personal
  photo}. Images which do not fit into any of these are labeled as
\textsl{other}. This includes images with inspirational quotes and
still shots of movies or TV shows.

\newpage

\paragraph{Image Content:}
To analyze image content we use the following prompt with GPT-4o to
extract a list of words describing the objects and scenes in each
image:

\say{Please analyze the attached image and list the contents of
  the image using 1-2 word phrases. If the image contains text, only
  list 'text', do not repeat the text contained in the image. Output the
  list one content per line, with no other information but the 1-2 word
  description.'}

We use GPT-4o for its combination of ease of use,
quality of output, and low cost. We validate the results by manually
checking the output for 100 randomly selected images. Of 673 content
words only 10 were false positives, 98.5\% precision. The top 50 most
common content words for each study can be seen in Table
\ref{tab:top_50}.

\begingroup
\centering\small
\begin{tabular}{lrlrlr}
  \toprule
  \multicolumn{2}{c}{Creation} & \multicolumn{2}{c}{Donation} & \multicolumn{2}{c}{Recent}        \\
  \cmidrule(r){1-2}\cmidrule(rl){3-4}\cmidrule(l){5-6}
  text & 496 & text& 627  & text &115  \\
  background & 140 & tree& 125  & tree& 27  \\
  tree & 137  & background &108 & woman& 26  \\
  sky& 128  & woman &105  & man& 23  \\
  grass& 127  & sky &98   & sky& 23  \\
  building& 100 & person &94  & background &21 \\
  water &97  & people &93  & person &20  \\
  people &94  & grass &92  & grass &18  \\
  woman& 77  & man& 89   & smile &16  \\
  car& 77   & smile &84  & building &16 \\
  light& 72  & light &74  & people& 15  \\
  blue &71  & building &72 & dog &14  \\
  cloud &71  & blue &63  & sunglasses &14 \\
  black& 69  & water &62  & shirt &14  \\
  wall &67  & table &59  & wall &13  \\
  man &64   & crowd& 53  & flower &13  \\
  green &64  & red &53   & water& 13  \\
  table& 61  & number &52  & table &12  \\
  red &61   & car& 50   & blue &12  \\
  smile& 54  & wall& 49  & chair &11  \\
  person& 53  & cloud &47  & cloud &11  \\
  chair &52  & flower &46  & sofa &9  \\
  street &50  & child &42  & crowd &9  \\
  White &48  & clothing &42 & light &9  \\
  sign& 46  & shirt &39  & red& 9   \\
  rock& 44  & black &39  & cat& 8   \\
  floor &43  & hand &38  & floor& 8  \\
  hand &43  & sign &37  & clothing& 8 \\
  flower &37  & street& 37  & pavement& 7 \\
  child &37  & floor &37  & mountain &7 \\
  face& 36  & green &36  & male\_child&6 \\
  hair &36  & expression &36 & hand &6  \\
  expression& 35 & drink &36  & rock &6  \\
  flag &35  & dog &36   & hat& 6   \\
  dog &34   & hair &34  & sunset& 6  \\
  shirt &33  & rock &34  & number &6  \\
  leaf &33  & chair &32  & drink &6  \\
  plant &33  & White &30  & orange &5  \\
  yellow &32  & decoration &29 & child& 5  \\
  clothing &32 & glass &26  & spectacles &5 \\
  drink& 31  & smoke &25  & black &5  \\
  eyes &30  & bag &25   & top& 5   \\
  hat& 29   & food &25  & hair& 5  \\
  suit &29  & dress &25  & two &5   \\
  hands& 28  & blanket& 25  & outdoor& 5  \\
  window& 28  & window& 25  & window& 5  \\
  field& 27  & plant &24  & White &5  \\
  beach& 27  & player& 24  & beach& 5  \\
  crowd &26  & orange &24  & colorful& 5 \\
  bag &25   & suit &24  & shadow &5  \\
  \bottomrule
\end{tabular}
\captionof{table}{\label{tab:top_50} Top 50 most common image content words for each approach, generated by GPT-4o.}
\endgroup

\newpage 

\subsubsection{Examples of posts}
\label{app:example_posts}
Here we look at specific examples of posts which highlight the differences described in Section \ref{sec:analysis}.

\begin{figure}[t]
  \centering
  \begin{subfigure}[t]{0.45\columnwidth}
    \centering
    \fbox{\includegraphics[width=\linewidth]{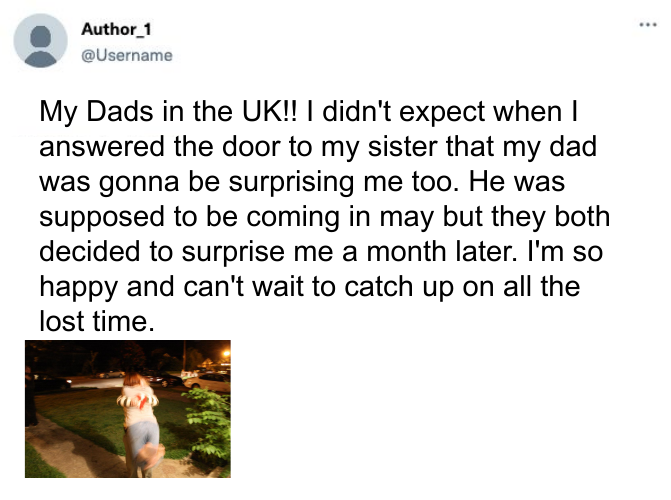}}
    \caption{\creation\ post labeled as surprise.}

  \end{subfigure}
  \hfill
  \begin{subfigure}[t]{0.45\columnwidth}
    \centering
    \fbox{\includegraphics[width=\linewidth]{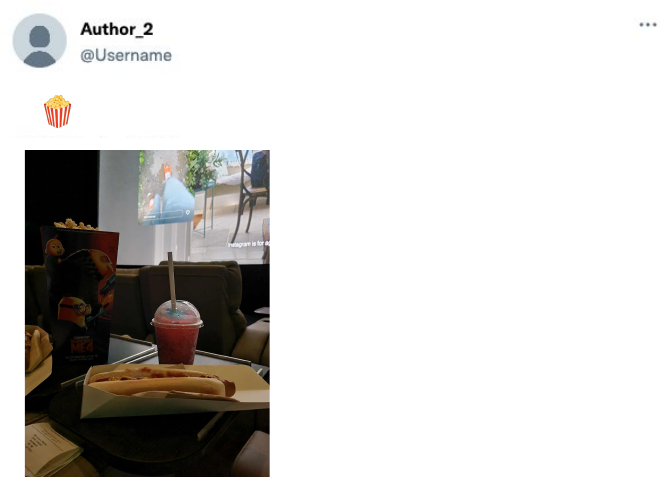}}
    \caption{\recent post labeled as joy.}

  \end{subfigure}
  \caption{Comparison of posts submitted for \creation\ and \recent. \creation\ gives a long detailed explanation of
    the event, including descriptions of their emotions, 
    while \recent uses only a single emoji to describe a similar event, spending time with their dad.}
  \label{Figure:Example_posts_length}
\end{figure}

Figure \ref{Figure:Example_posts_length} shows how length and detail of description vary by collection strategy. In post 
(a) from \creation\, you can see how the author explicitly describes both the event and the emotion they experienced. In post 
(b) from \recent\, the author uses only a single emoji to describe a similar event, spending time with their Dad. The event 
can only be gleaned from
the additional annotation in which the author describes it as \say{I felt happy because I was spending time with my dad}.

\begin{figure}[t]
  \centering
  \begin{subfigure}[t]{0.45\columnwidth}
    \centering
    \fbox{\includegraphics[width=\linewidth]{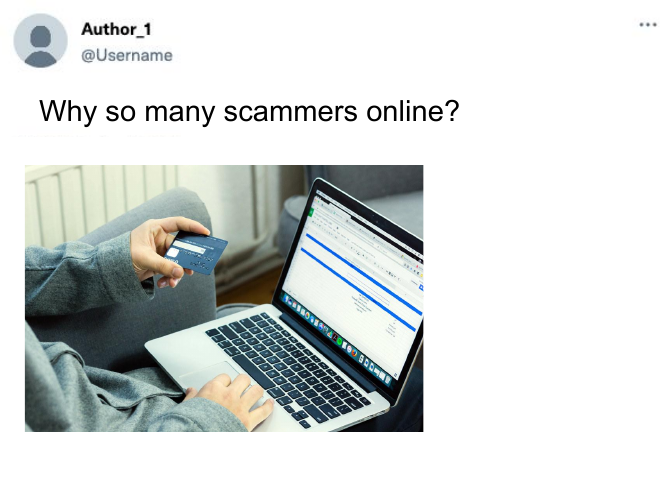}}
    \caption{\creation\ post labeled as anger.}

  \end{subfigure}
  \hfill
  \begin{subfigure}[t]{0.45\columnwidth}
    \centering
    \fbox{\includegraphics[width=\linewidth]{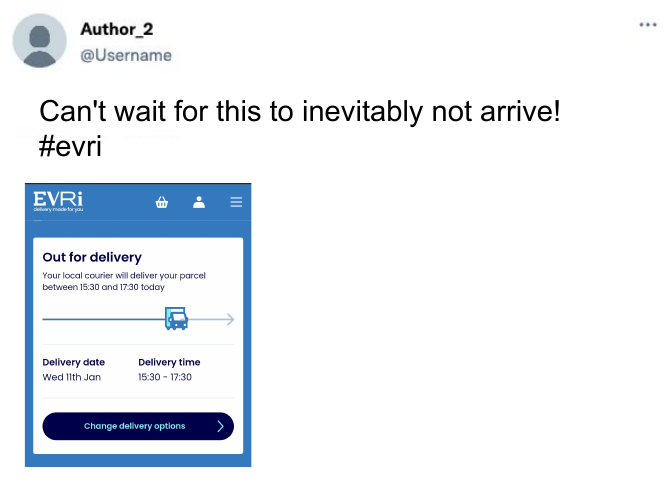}}
    \caption{\donation\ post labeled as anger.}

  \end{subfigure}
  \caption{Comparison of posts submitted for \creation\ and \donation. \creation\ uses a stock photo, 
    while \donation\ demonstrates a reaction post in which the author uses a screenshot of an event they 
    experienced to illustrate their emotion.}
  \label{Figure:Example_posts_image_type}
\end{figure}

Figure \ref{Figure:Example_posts_image_type} highlights how image types can affect posts: post (a) uses a stock photo to stand in as a representation of the 
event which triggered anger in them. The specific event is not detailed and the generic image does not help give any more clues 
to the reader. The post from \donation, however, uses a screenshot that directly references the event which triggered the author'
anger. These reaction posts are not possible in \creation, since the participants only have access to the image database, and not
their own personal images. This is also another example of \creation posts being about general events, where \donation tend to be 
about specific events.

\begin{figure}[t]
  \centering
  \begin{subfigure}[t]{0.45\columnwidth}
    \centering
    \fbox{\includegraphics[width=\linewidth]{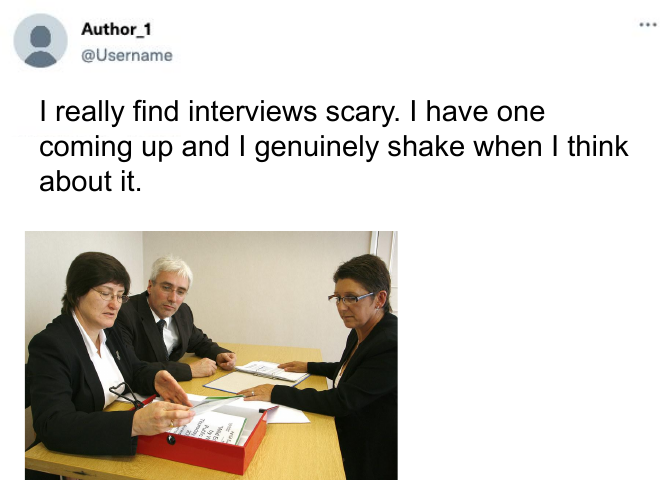}}
    \caption{\creation\ post labeled as fear.}

  \end{subfigure}
  \hfill
  \begin{subfigure}[t]{0.45\columnwidth}
    \centering
    \fbox{\includegraphics[width=\linewidth]{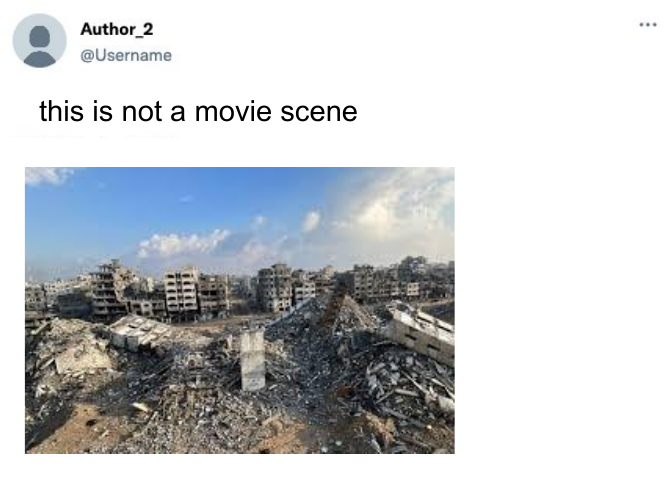}}
    \caption{\donation\ post labeled as fear.}

  \end{subfigure}
  \caption{Comparison of posts submitted for \creation\ and \donation. \creation's image conveys no emotion and relies entirely, 
    on the text to convey emotion, while \donation's text is emotion neutral and relies on the image to convey the experience of fear. 
    experienced to illustrate their emotion.}
  \label{Figure:Example_posts_relation}
\end{figure}

Figure \ref{Figure:Example_posts_relation} illustrates how posts can rely more on the text or the image to 
convey emotion. In post (a), a \creation post, the image conveys almost no emotion, while the text explicitly 
describes the author's emotion. Post (b), a \donation post, on the other hand, conveys no emotion in the text, 
relying on the image to convey the author's emotion of fear.

\begin{figure}[t]
  \centering
  \begin{subfigure}[t]{0.45\columnwidth}
    \centering
    \fbox{\includegraphics[width=\linewidth]{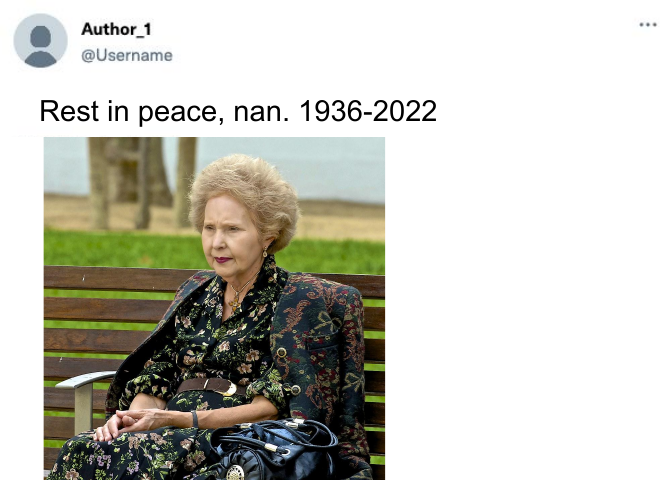}}
    \caption{\creation\ post labeled as sadness.}

  \end{subfigure}
  \hfill
  \begin{subfigure}[t]{0.45\columnwidth}
    \centering
    \fbox{\includegraphics[width=\linewidth]{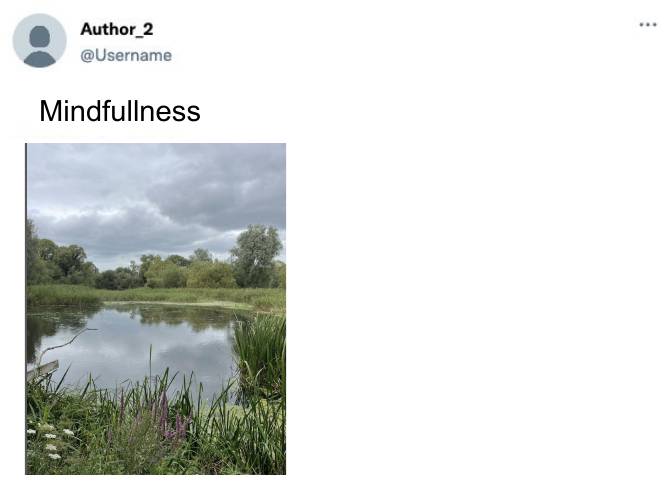}}
    \caption{\donation\ post labeled as sadness.}

  \end{subfigure}
  \caption{Comparison of posts submitted for \creation\ and \donation. \creation\ is a prototypical event for sadness, 
    while \donation\ is a more general sense of sadness.}
  \label{Figure:proto_event_posts}
\end{figure}

In Figure \ref{Figure:proto_event_posts} we can see that post (a) was inspired by a prototypical event for sadness, the death of a 
loved one. Post (b) is less obvious what event inspired the author to post, or indeed which emotion they are expressing. The author
describes the event as "I was feeling upset that day and so was on a walk around a local nature reserve to unwind". In this case
the event is more difficult to define.

\subsubsection{Platform distribution between approaches}
\label{app:platforms}

\begin{figure}[h]
  \centering
  \includegraphics[width=0.5\columnwidth]{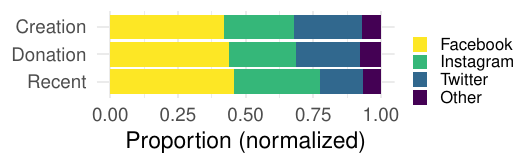}
  \caption{\protect Distribution of post platforms under the \creation, \donation, and \recent approaches. As multiple responses were allowed in \creation, proportions are normalized such that they sum to one.}
  \label{fig:platforms}
\end{figure}

Figure \ref{fig:platforms} shows the distribution of posts between social media platforms. Participants in \donation and \recent were asked which platform their donated post was originally posted on. Participants in \creation were asked which platform(s) they would have posted their study-created post on. 

\newpage

\subsection{Model Details}
\label{app:modeldetails}

All models are trained using four NVIDIA L40 GPUs. Each supervised model is fine-tuned five times, using the exact same setup and
environment. For zero-shot models, every prompt is run five times per model. The average and range of scores is reported. All code
can be found in the supplementary materials.\footnote{\url{https://www.uni-bamberg.de/en/nlproc/projects/item/}}
\subsubsection{Fine-tuned models}
\paragraph{Text-only:}
We use RoBERTa-base pretrained model for both tokenizer and model, via the transformers Python package.\footnote{\url{https://huggingface.co/transformers/v2.9.1/model_doc/roberta.html}}
All text-only models are trained using 10 epochs, 1e-5 training rate, 16 batch size, 0.01 weight decay, and early stopping.

\paragraph{Vision-only:}
We fine-tune the vit-base-patch16-224 model using a combination of transformers and torchvision Python packages.\footnote{\url{https://huggingface.co/google/vit-base-patch16-224}}
Images are preproccessed by resizing them to 224x224.
All image-only models are trained using 5 epochs, 5e-5 learning rate, 32 batch size, cross entropy loss, and early stopping.

\paragraph{Text+Vision:}
We fine-tune the clip-vit-base-patch32 model using a combination of transformers and torchvision Python packages.\footnote{\url{https://huggingface.co/openai/clip-vit-base-patch32}}
Both images and text are encoded via the CLIP processor, and then fused using a simple concatenation method. All text+vision models are trained using 10 epochs,
1e-5 learning rate, batch size 8, cross entropy loss, and early stopping. 

\subsubsection{Zero-shot models}

We use six multimodal models made available through the Ollama Python package\footnote{\url{https://pypi.org/project/ollama/0.4.6/}}. These are 
llama3.2-vision\footnote{\url{https://ollama.com/library/llama3.2-vision}} (11b, ID 085a1fdae525), 
minicpm-v\footnote{\url{https://ollama.com/library/minicpm-v}} (8b, v2.6, ID c92bfad01205), 
llava\footnote{\url{https://ollama.com/library/llava}} (7b, v1.6, ID 8dd30f6b0cb1), 
llava-llama3\footnote{\url{https://ollama.com/library/llava-llama3}} (8b, ID 44c161b1f465), 
bakllava\footnote{\url{https://ollama.com/library/bakllava}} (7b, ID 3dd68bd4447c), 
and llava-phi3\footnote{\url{https://ollama.com/library/llava-phi3}} (3.8b, ID c7edd7b87593). 
Bakllava and llava-phi3 returned nearly entirely unparseable results and  were hence  eliminated from further analyis.

\begin{table*}
  \centering\sffamily\small\scalefont{0.85}
  \begin{tabularx}{\linewidth}{p{10mm} XXX}
    \toprule
    Section & Text & Vision & Text + Vision \\
    \cmidrule(r){1-1}\cmidrule(rl){2-2}\cmidrule(rl){3-3}\cmidrule(rl){4-4}
    Task \newline Descr.
            &Which emotion does the following text from a social media post convey most strongly?
                   &Which emotion does the following image from a social media post convey most strongly?
                            &Which emotion does the following text and image from a social media post convey most strongly?
    \\    \cmidrule(r){1-1}\cmidrule(rl){2-2}\cmidrule(rl){3-3}\cmidrule(rl){4-4}
    Labels
            &Please choose one of the set [anger, disgust, fear, joy, sadness, surprise].
                   &Please choose one of the set [anger, disgust, fear, joy, sadness, surprise].
                            &Please choose one of the set [anger, disgust, fear, joy, sadness, surprise].
    \\    \cmidrule(r){1-1}\cmidrule(rl){2-2}\cmidrule(rl){3-3}\cmidrule(rl){4-4}
    Format \newline Instr.
            &Only provide a single word indicating the emotion. Do not provide explanation or analysis. Only provide plain text.
                   &Only provide a single word indicating the emotion. Do not provide explanation or analysis. Only provide plain text.
                            &Only provide a single word indicating the emotion. Do not provide explanation or analysis. Only provide plain text.
    \\    \cmidrule(r){1-1}\cmidrule(rl){2-2}\cmidrule(rl){3-3}\cmidrule(rl){4-4}
    Data \newline Input
            &Post text: \{text\}
                   &
                            &Post text: \{text\}
    \\    \cmidrule(r){1-1}\cmidrule(rl){2-2}\cmidrule(rl){3-3}\cmidrule(rl){4-4}
    Image
            &
                   & this\_image.\{image type\}
                            & this\_image.\{image type\}
    \\
    \bottomrule
  \end{tabularx}
  \caption{Prompts for text, vision, and text + vision modalities. Variables are typeset in \{curly
    brackets\}. Image filenames were changed to a default name before sending to the LLM to 
    avoid the model being able to gain additional information.}
  \label{tab:prompts}
\end{table*}

We perform basic cleaning on responses (converting to lowercase,
removing punctuation and whitespace, removing the word ``emotion''). We
continue to prompt the model until we received five valid emotion
predictions, with ``valid'' defined as equalling one of the six provided
emotion options after cleaning. We prompt models up to a maximum of
30 times for each post. In the small number of cases in which we did
not receive five valid predictions in 30 tries, we count missing
responses as incorrect for the purposes of performance statistic calculations.

\newpage
\subsection{Additional Modeling Results}
\label{app:modeling}

\begin{table*}[h!]
  \centering\small
  \begin{tabular}{llc cc cc cc cc}
    \toprule
    &&& \multicolumn{2}{c}{llama3.2-vision} & \multicolumn{2}{c}{minicpm-v} & \multicolumn{2}{c}{llava}
    & \multicolumn{2}{c}{llava-llama3}  \\ \cmidrule(lr){3-5}\cmidrule(lr){6-7}\cmidrule(lr){8-9}\cmidrule{10-11}
    && Modality & \F & Range & \F & Range & \F & Range & \F & Range \\ 
    \cmidrule(lr){3-3}\cmidrule(lr){4-5}\cmidrule(lr){6-7}\cmidrule(lr){8-9}\cmidrule{10-11}
    \multirow{6}{*}{\rotatebox{90}{Test}} 
    &\multirow{3}{*}{\rotatebox{90}{Creatn}}
      &Vision   & .24 & .20--.30 & .21 & .20--.24 & .25 & .21--.31 & .25 & .21--.33 \\
    &&Text   & .61 & .60--.63 & .58 & .54--.65 & .55 & .52--.58 & .49 & .41--.55 \\
    &&T+V & .53 & .50--.57 & .56 & .52--.60 & .44 & .42--.48 & .43 & .37--.49 \\
    \cmidrule(lr){2-3}\cmidrule(lr){4-5}\cmidrule(lr){6-7}\cmidrule(lr){8-9}\cmidrule{10-11}
    &\multirow{3}{*}{\rotatebox{90}{Donatn}}
      &Vision   & .21 & .17--.25 & .25 & .21--.30 & .23 & .19--.28 & .19 & .14--.24 \\
    &&Text   & .45 & .41--.48 & .45 & .42--.49 & .42 & .39--.45 & .44 & .41--.47 \\
    &&T+V & .39 & .35--.43 & .43 & .41--.45 & .33 & .28--.38 & .26 & .18--.30 \\
    \bottomrule
  \end{tabular}    
  \caption{Macro \F of multimodal zero-shot models for predicting
    emotion on \creation and \donation posts using post text (T),
    post image (V), and both text and image (T+V).}
  \label{tab:zeroshotapp}
\end{table*}

\begin{table*}
  \centering\small
  \begin{tabular}{llll ccc ccc ccc}
    \toprule
    &&&&\multicolumn{6}{c}{Training Data}\\
    \cmidrule(lr){5-10}
    &&&&\multicolumn{3}{c}{\creation} & \multicolumn{3}{c}{\donation} & \multicolumn{3}{c}{ Zero-shot}  \\
    \cmidrule(lr){5-7}\cmidrule(lr){8-10}\cmidrule(lr){11-13}
    &&Modality & Emotion & \F & P & R & \F & P & R & \F & P & R \\
    \cmidrule(r){3-4}\cmidrule(lr){5-7}\cmidrule(lr){8-10}\cmidrule(l){11-13}
    \multirow{42}{*}{\rotatebox{90}{Test}}&\multirow{18}{*}{\rotatebox{90}{Creation}}&
                                                                                       Vision   & anger    & .25 & .27 & .21 & .17 & .17 & .14 & .01\textsuperscript{1} & .10\textsuperscript{1} & .01\textsuperscript{1} \\
    &&Vision   & disgust  & .16 & .07 & .16 & .15 & .10 & .16 & .19\textsuperscript{1} & .30\textsuperscript{1} & .14\textsuperscript{1} \\
    &&Vision   & fear     & .12 & .12 & .12 & .18 & .10 & .25 & .14\textsuperscript{1} & .31\textsuperscript{1} & .09\textsuperscript{1} \\
    &&Vision   & joy      & .11 & .20 & .12 & .06 & .09 & .05 & .39\textsuperscript{1} & .26\textsuperscript{1} & .78\textsuperscript{1} \\
    &&Vision   & sadness  & .23 & .29 & .25 & .20 & .14 & .19 & .26\textsuperscript{1} & .26\textsuperscript{1} & .26\textsuperscript{1} \\
    &&Vision   & surprise & .22 & .23 & .24 & .17 & .14 & .17 & .18\textsuperscript{1} & .17\textsuperscript{1} & .20\textsuperscript{1} \\
    \cmidrule(r){3-4}\cmidrule(lr){5-7}\cmidrule(lr){8-10}\cmidrule(l){11-13}
    &&Text   & anger    & .46 & .42 & .40 & .47 & .50 & .49 & .65\textsuperscript{1} & .60\textsuperscript{1} & .71\textsuperscript{1} \\
    &&Text   & disgust  & .57 & .41 & .67 & .45 & .31 & .60 & .55\textsuperscript{1} & .76\textsuperscript{1} & .43\textsuperscript{1} \\
    &&Text   & fear     & .50 & .52 & .49 & .34 & .33 & .29 & .59\textsuperscript{1} & .89\textsuperscript{1} & .44\textsuperscript{1} \\
    &&Text   & joy      & .73 & .67 & .77 & .70 & .58 & .65 & .76\textsuperscript{1} & .63\textsuperscript{1} & .95\textsuperscript{1} \\
    &&Text   & sadness  & .61 & .57 & .64 & .57 & .41 & .66 & .57\textsuperscript{1} & .44\textsuperscript{1} & .83\textsuperscript{1} \\
    &&Text   & surprise & .57 & .79 & .50 & .28 & .17 & .22 & .20\textsuperscript{1} & .59\textsuperscript{1} & .12\textsuperscript{1} \\
    \cmidrule(r){3-4}\cmidrule(lr){5-7}\cmidrule(lr){8-10}\cmidrule(l){11-13}
    &&Text + Vision & anger    & .55 & .62 & .51 & .50 & .45 & .50 & .58\textsuperscript{2} & .51\textsuperscript{2} & .70\textsuperscript{2} \\
    &&Text + Vision & disgust  & .55 & .55 & .61 & .53 & .42 & .58 & .53\textsuperscript{2} & .66\textsuperscript{2} & .44\textsuperscript{2} \\
    &&Text + Vision & fear     & .57 & .52 & .60 & .56 & .65 & .54 & .56\textsuperscript{2} & .75\textsuperscript{2} & .45\textsuperscript{2} \\
    &&Text + Vision & joy      & .79 & .74 & .82 & .77 & .83 & .78 & .71\textsuperscript{2} & .56\textsuperscript{2} & .97\textsuperscript{2} \\
    &&Text + Vision & sadness  & .65 & .63 & .66 & .62 & .59 & .65 & .60\textsuperscript{2} & .54\textsuperscript{2} & .68\textsuperscript{2} \\
    &&Text + Vision & surprise & .59 & .77 & .50 & .61 & .71 & .55 & .20\textsuperscript{2} & .39\textsuperscript{2} & .13\textsuperscript{2} \\
    \cmidrule(l){2-13}
    &\multirow{18}{*}{\rotatebox{90}{Donation}}&
                                                 Vision   & anger    & .13 & .28 & .10 & .22 & .12 & .21 & .34\textsuperscript{3} & .34\textsuperscript{3} & .34\textsuperscript{3} \\
    &&Vision   & disgust  & .14 & .25 & .14 & .16 & .27 & .14 & .03\textsuperscript{3} & .08\textsuperscript{3} & .02\textsuperscript{3} \\
    &&Vision   & fear     & .19 & .24 & .18 & .23 & .15 & .28 & .07\textsuperscript{3} & .10\textsuperscript{3} & .06\textsuperscript{3} \\
    &&Vision   & joy      & .11 & .20 & .12 & .03 & .06 & .03 & .36\textsuperscript{3} & .23\textsuperscript{3} & .92\textsuperscript{3} \\
    &&Vision   & sadness  & .21 & .15 & .19 & .26 & .22 & .26 & .00\textsuperscript{3} & .00\textsuperscript{3} & .00\textsuperscript{3} \\
    &&Vision   & surprise & .27 & .25 & .33 & .19 & .13 & .21 & .10\textsuperscript{3} & .22\textsuperscript{3} & .06\textsuperscript{3} \\
    \cmidrule(r){3-4}\cmidrule(lr){5-7}\cmidrule(lr){8-10}\cmidrule(l){11-13}
    &&Text   & anger    & .36 & .41 & .31 & .47 & .43 & .44 & .54\textsuperscript{1} & .57\textsuperscript{1} & .52\textsuperscript{1} \\
    &&Text   & disgust  & .48 & .42 & .50 & .34 & .27 & .41 & .30\textsuperscript{1} & .44\textsuperscript{1} & .22\textsuperscript{1} \\
    &&Text   & fear     & .35 & .33 & .36 & .18 & .10 & .13 & .40\textsuperscript{1} & .68\textsuperscript{1} & .29\textsuperscript{1} \\
    &&Text   & joy      & .55 & .38 & .81 & .61 & .48 & .74 & .53\textsuperscript{1} & .40\textsuperscript{1} & .79\textsuperscript{1} \\
    &&Text   & sadness  & .43 & .50 & .39 & .45 & .39 & .44 & .39\textsuperscript{1} & .30\textsuperscript{1} & .54\textsuperscript{1} \\
    &&Text   & surprise & .20 & .30 & .14 & .33 & .31 & .32 & .27\textsuperscript{1} & .53\textsuperscript{1} & .18\textsuperscript{1} \\
    \cmidrule(r){3-4}\cmidrule(lr){5-7}\cmidrule(lr){8-10}\cmidrule(l){11-13}
    &&Text + Vision & anger    & .34 & .37 & .34 & .33 & .25 & .34 & .51\textsuperscript{2} & .48\textsuperscript{2} & .54\textsuperscript{2} \\
    &&Text + Vision & disgust  & .29 & .32 & .27 & .28 & .24 & .28 & .27\textsuperscript{2} & .41\textsuperscript{2} & .21\textsuperscript{2} \\
    &&Text + Vision & fear     & .37 & .36 & .38 & .35 & .44 & .34 & .33\textsuperscript{2} & .51\textsuperscript{2} & .25\textsuperscript{2} \\
    &&Text + Vision & joy      & .57 & .52 & .68 & .57 & .47 & .67 & .50\textsuperscript{2} & .35\textsuperscript{2} & .87\textsuperscript{2} \\
    &&Text + Vision & sadness  & .42 & .48 & .42 & .38 & .45 & .38 & .44\textsuperscript{2} & .43\textsuperscript{2} & .46\textsuperscript{2} \\
    &&Text + Vision & surprise & .36 & .38 & .32 & .34 & .32 & .30 & .25\textsuperscript{2} & .53\textsuperscript{2} & .17\textsuperscript{2} \\
    \bottomrule
  \end{tabular}    
  \caption{\label{app:modelingapdx}Model predictive performance by
    emotion. Results shown are averaged over 5 runs. Zero-shot models
    are chosen for each data-test set combination based on overall performance on development data, and are the same here a reported in the main
    text. \textsuperscript{1}llama3.2-vision. \textsuperscript{2}minicpm-v. \textsuperscript{3}llava-llama3.}
\end{table*}

\onecolumn
\newcommand{\mpy}{\multicolumn{2}{l}{\hspace{7mm}------}}
\begin{small}
\begin{longtable}{l rl rl rl}
\caption{Coefficients for logistic regressions predicting correct model emotion
classification, in log odds units. Numeric and ordinal variables are scaled to range from 0--1. Reference
levels -- which serve as baselines for each variable -- are indicated with hyphens in the table. They are
chosen according to \citet{johfre2021reconsidering}, and are in general either the lowest category
conceptually (in the case of inherently ordered variables like education) or lowest category numerically in terms
of their effect on the dependent variable (in the case of unordered variables like emotion). Clustered
standard errors (reported in parentheses) are used to account for modeling multiple predictions of
the same posts. \textsuperscript{+}p<.1. \textsuperscript{*}p<.05. \textsuperscript{**}p<.01.
\textsuperscript{***} p<.001.}\\
\label{app:regressiontab}\\
\toprule
  & \multicolumn{2}{c}{CLIP: creation-trained} & \multicolumn{2}{c}{CLIP: donation-trained} & \multicolumn{2}{c}{minicpm-v} \\
  \cmidrule(lr){2-3}\cmidrule(lr){4-5}\cmidrule(l){6-7}
  & Coef & SE & Coef & SE & Coef & SE \\
  \cmidrule(r){1-1}\cmidrule(lr){2-3}\cmidrule(lr){4-5}\cmidrule(l){6-7}
  \endhead
  \\
  \multicolumn{3}{r@{}}{continued \ldots}\\
  \bottomrule
  \endfoot
  \bottomrule
  \endlastfoot
  
  (Intercept) & $-$1.27 & (0.98) & $-$0.49 & (0.94) & $-$2.85 & (1.49)+ \\
\textbf{Emotion} &&&&&& \\
\nt{}surprise & \mpy & \mpy & \mpy \\
\nt{}anger & 0.42 & (0.54) & $-$0.19 & (0.54) & 2.26 & (0.70)** \\
\nt{}disgust & 0.64 & (0.51) & 0.26 & (0.48) & 1.02 & (0.67) \\
\nt{}fear & 0.85 & (0.53) & 0.40 & (0.53) & 0.87 & (0.67) \\
\nt{}joy & 1.79 & (0.45)*** & 1.76 & (0.43)*** & 5.16 & (0.89)*** \\
\nt{}sadness & 1.05 & (0.56)+ & 0.47 & (0.55) & 2.22 & (0.74)** \\
\textbf{Study} &&&&&& \\
\nt{}creation & \mpy & \mpy & \mpy \\
\nt{}donation & $-$0.53 & (0.26)* & $-$0.51 & (0.25)* & $-$0.59 & (0.37) \\
\textbf{Text-image relation} &&&&&& \\
\nt{}Text\,describe\,image & $-$0.30 & (0.37) & 0.13 & (0.34) & 0.31 & (0.50) \\
\nt{}Text\,understand\,image & $-$0.28 & (0.36) & $-$0.36 & (0.34) & $-$0.49 & (0.46) \\
\nt{}Image\,understand\,text & $-$0.50 & (0.36) & $-$0.46 & (0.33) & $-$1.02 & (0.45)* \\
\nt{}Image\,conveys\,emotion & $-$0.14 & (0.43) & $-$0.65 & (0.47) & 0.08 & (0.53) \\
\nt{}Text\,conveys\,emotion & 0.41 & (0.47) & 0.67 & (0.47) & $-$0.03 & (0.51) \\
\textbf{Emotion appraisals} &&&&&& \\
\nt{}Familiarity & 0.18 & (0.32) & $-$0.01 & (0.32) & 0.17 & (0.39) \\
\nt{}Predictability & $-$0.20 & (0.40) & 0.07 & (0.40) & $-$0.74 & (0.58) \\
\nt{}Pleasantness & 0.30 & (0.77) & $-$0.20 & (0.75) & $-$0.49 & (1.04) \\
\nt{}Unpleasantness & 0.34 & (0.66) & 0.28 & (0.69) & 1.07 & (0.78) \\
\nt{}Goalrelevance & 0.09 & (0.44) & 0.00 & (0.44) & $-$0.14 & (0.58) \\
\nt{}Ownresponsibility & $-$0.04 & (0.44) & 0.01 & (0.41) & 0.10 & (0.54) \\
\nt{}Anticip.conseq & 0.24 & (0.44) & 0.09 & (0.44) & 0.69 & (0.59) \\
\nt{}Goalsupport & 0.27 & (0.49) & $-$0.01 & (0.50) & 0.28 & (0.72) \\
\nt{}Own\,control & $-$0.08 & (0.49) & 0.05 & (0.47) & $-$0.38 & (0.62) \\
\nt{}Others\,control & 0.25 & (0.33) & $-$0.12 & (0.32) & $-$0.07 & (0.42) \\
\nt{}Accept\,conseq. & $-$0.35 & (0.37) & $-$0.48 & (0.39) & $-$0.52 & (0.50) \\
\nt{}Internal\,standards & $-$0.73 & (0.44)+ & $-$0.70 & (0.42)+ & $-$0.99 & (0.51)+ \\
\nt{}Attention & 0.41 & (0.46) & 0.60 & (0.46) & 0.52 & (0.56) \\
\nt{}Not\,consider & 0.18 & (0.44) & 0.09 & (0.41) & 0.62 & (0.52) \\
\nt{}Effort & 0.09 & (0.45) & $-$0.29 & (0.46) & $-$0.58 & (0.55) \\
\textbf{Participant gender identity} &&&&&& \\
\nt{}Nonman & \mpy & \mpy & \mpy \\
\nt{}Man & 0.40 & (0.27) & 0.68 & (0.27)* & 0.63 & (0.36)+ \\
\nt{}Age & 1.43 & (0.77)+ & 1.22 & (0.73)+ & 0.70 & (0.96) \\
\nt{}Participant ethnicity &&&&&& \\
\nt{}Non-European or multiethnic & \mpy & \mpy & \mpy \\
\nt{}European & 0.16 & (0.34) & $-$0.03 & (0.32) & 0.08 & (0.39) \\
    \pagebreak
\textbf{Participant education} &&&&&& \\
\nt{}No\,college & \mpy & \mpy & \mpy \\
\nt{}B.Sc. & 0.20 & (0.28) & 0.10 & (0.28) & 0.07 & (0.33) \\
\nt{}Masters\,or\,more & $-$0.33 & (0.35) & 0.15 & (0.35) & $-$1.03 & (0.42)* \\
\textbf{Participant employment status} &&&&&& \\
\nt{}Not full-time & \mpy & \mpy & \mpy \\
\nt{}Full-Time & $-$0.06 & (0.26) & $-$0.06 & (0.26) & $-$0.01 & (0.33) \\
\textbf{Participant is a student} &&&&&& \\
\nt{}No & \mpy & \mpy & \mpy \\
\nt{}Yes & 0.02 & (0.32) & $-$0.39 & (0.30) & 0.21 & (0.43) \\
\textbf{Participant social media frequency} &&&&&& \\
\nt{}Use social media & $-$0.04 & (0.44) & 0.06 & (0.46) & 1.16 & (0.86) \\
\nt{}Post on social media & $-$0.31 & (0.42) & $-$0.21 & (0.38) & $-$0.03 & (0.52) \\
\textbf{Participant preferred platform} &&&&&& \\
\nt{}Facebook & \mpy & \mpy & \mpy \\
\nt{}Instagram & 0.63 & (0.33)+ & 0.50 & (0.31) & 0.67 & (0.45) \\
\nt{}X (Twitter) & 0.30 & (0.40) & $-$0.17 & (0.38) & 0.45 & (0.54) \\
\nt{}Other platform & 0.33 & (0.41) & 0.19 & (0.38) & 0.31 & (0.50) \\
\textbf{Image style} &&&&&& \\
\nt{}Personal Photo & \mpy & \mpy & \mpy \\
\nt{}Pro Photo & 0.10 & (0.31) & 0.02 & (0.33) & $-$0.02 & (0.43) \\
\nt{}Other image style & $-$0.33 & (0.36) & $-$0.28 & (0.36) & 0.02 & (0.42) \\
\textbf{Post text length} & $-$0.11 & (1.12) & $-$0.82 & (1.09) & 0.63 & (2.31) \\
\textbf{Duration and intensity} &&&&&& \\
\nt{}Event duration & $-$0.94 & (0.64) & $-$0.55 & (0.63) & $-$0.11 & (0.75) \\
\nt{}Event intensity & 0.14 & (0.78) & $-$0.08 & (0.80) & 1.59 & (1.06) \\
\nt{}Emotion duration & 0.37 & (0.75) & 0.12 & (0.77) & $-$0.58 & (0.84) \\
\nt{}Emotion intensity & $-$0.04 & (0.73) & 0.38 & (0.72) & $-$1.37 & (1.06) \\
\nt{}Num.Obs. & 1365 &  & 1365 &  & 1362 &  \\
\nt{}RMSE & 0.45 &  & 0.45 &  & 0.39 &  \\

\end{longtable}
\end{small}

\end{document}